\documentclass[lettersize,journal]{IEEEtran}
\usepackage{amsmath,amsfonts,amssymb}
\usepackage{algorithm}
\usepackage{array}
\usepackage{textcomp}
\usepackage{stfloats}
\usepackage{url}
\usepackage{verbatim}
\usepackage{graphicx}
\usepackage{cite}

\usepackage{makecell}
\usepackage{algpseudocode}
\usepackage{multirow} 
\usepackage{caption}
\usepackage{float} 
\usepackage{subcaption}
\usepackage{color}
\usepackage{booktabs}
\usepackage{mathabx}
\usepackage{hyperref}

\newtheorem{theorem}{Theorem}
\newtheorem{proposition}[theorem]{Proposition}
\newtheorem{corollary}[theorem]{Corollary}

\begin{document}

\title{Improving Uncertainty Quantification of Variance Networks by Tree-Structured Learning}

\author{Wenxuan MA, Xing YAN, and Kun ZHANG%~\IEEEmembership{Staff,~IEEE,}
        % <-this % stops a space
\thanks{The authors are with the Institute of Statistics and Big Data, Renmin University of China (e-mail: \{mawenxuan,xingyan,kunzhang\}@ruc.edu.cn).}
\thanks{Corresponding author: Xing YAN.}
%\thanks{This paper was produced by the IEEE Publication Technology Group. They are in Piscataway, NJ.}% <-this % stops a space
%\thanks{Manuscript received April 19, 2021; revised August 16, 2021.}
}

\markboth{Improving Uncertainty Quantification of Variance Networks by Tree-Structured Learning}%
{Shell \MakeLowercase{\textit{et al.}}: A Sample Article Using IEEEtran.cls for IEEE Journals}


\maketitle

\begin{abstract}%   <- trailing '%' for backward compatibility of .sty file
        To improve the uncertainty quantification of variance networks, we propose a novel tree-structured local neural network model that partitions the feature space into multiple regions based on uncertainty heterogeneity. A tree is built upon giving the training data, whose leaf nodes represent different regions where region-specific neural networks are trained to predict both the mean and the variance for quantifying uncertainty. The proposed Uncertainty-Splitting Neural Regression Tree (USNRT) employs novel splitting criteria. At each node, a neural network is trained on the full data first, and a statistical test for the residuals is conducted to find the best split, corresponding to the two sub-regions with the most significant uncertainty heterogeneity between them. USNRT is computationally friendly because very few leaf nodes are sufficient and pruning is unnecessary. Furthermore, an ensemble version can be easily constructed to estimate the total uncertainty including the aleatory and epistemic. On extensive UCI datasets, USNRT or its ensemble shows superior performance compared to some recent popular methods for quantifying uncertainty with variances. Through comprehensive visualization and analysis, we uncover how USNRT works and show its merits, revealing that uncertainty heterogeneity does exist in many datasets and can be learned by USNRT.
        \end{abstract}
        
        
        \section{Introduction}
        
                Deep learning has achieved great successes in many fields or applications, such as computer vision, natural language processing,  protein structure prediction, and game playing \cite{lecun2015deep}. 
                The most attractive feature of deep learning is the remarkable prediction accuracy it can obtain.
                However, there are still serious concerns preventing deep learning from deploying extensively in real life, including their opacity in making predictions, their fragility in facing diverse inputs, and their over-confidence about the prediction results. Especially in risk-sensitive decision-makings, practitioners do not allow certain types of errors made by machines.
                Uncertainty quantification aims to alleviate those concerns partly by giving a confidence accompanied with each prediction result. Good uncertainty estimation can avoid over-confidence \cite{guo2017calibration,rahaman2021uncertainty} and reduce the risk of making unacceptable errors for decision-makers. Therefore, it is worthwhile to make efforts to produce good uncertainty estimation and only trust outcomes with high confidence.
        
        There are several types of uncertainty quantification methods for neural networks proposed recently. 
        A large body of literature focuses on outputting a variance as the uncertainty measure.
        Among these works, Bayesian methods \cite{blundell2015weight, hernandez2015probabilistic, gal2016dropout, gal2017concrete, maddox2019simple} use the estimated posterior density function (simplified as predictive variance) to quantify uncertainty, generally with Gaussian likelihood. The approximation computation will be needed to overcome the computational issue.
        %However, these models have high computational costs and depend on whether the prior distribution is correctly selected \cite{lakshminarayanan2017simple}. 
        Also with Gaussian assumption, some works \cite{lakshminarayanan2017simple, skafte2019reliable, zhao2020individual, cui2020calibrated} use neural network models to produce two outputs, the mean and the variance, for quantifying uncertainty. While one mean and one variance are used to estimate the aleatory only, multiple means and variances obtained from multiple networks are for the total uncertainty as the Bayesian methods. The recent Deep Evidential Regression \cite{amini2020deep} also belongs to this class.
        Another big class of works, the quantile regression-based models \cite{tagasovska2019single, romano2019conformalized, chung2021beyond}, quantify uncertainty by estimating a group of conditional quantiles to  form prediction intervals. However, in this paper, we focus on the methodology of variance networks only.
        
        \IEEEpubidadjcol
        
        A key observation made by us is that data exhibits uncertainty heterogeneity across different regions in the feature space, as illustrated in Fig. \ref{split_uncertainty} and \ref{Uncertainty_difference} in the experimental section. Thereby, discovering heterogeneity via partitioning the space is motivated.
        With the emergence of complex, diverse, and large-scale datasets, reducing the difficulty of modeling through space partitioning or data partitioning becomes necessary, making machine learning tasks (uncertainty quantification tasks in this paper) less challenging.
            In the past, some researchers used data partitioning as a pre-method \cite{fan2018rectangular,ge2019random} and sub-model was fitted within each region. Some region-specific models \cite{friedman1991multivariate,wang2012local, oiwa2014partition} and tree-based models \cite{ chaudhuri1994piecewise, zheng2019partitioning} were proposed with similar goals. As far as we know, there were very few works tackling the challenging uncertainty quantification tasks of deep learning with data heterogeneity and data partitioning considerations.
                
                In this paper, we propose a novel tree-based neural network model called Uncertainty-Splitting Neural Regression Tree (USNRT) which partitions the feature space recursively based on novel splitting criteria for discovering uncertainty heterogeneity. It gives the prediction and uncertainty estimation based on region-specific sub-models.
                For the splitting criteria, a neural network is trained first at each internal node and a statistical test is conducted on the residuals to help assess the uncertainty heterogeneity between the two sub-regions.
                In each leaf region, two region-specific neural networks are used to predict the mean and the variance respectively. Intuitively, this can reduce the difficulty of modeling compared to the full-space modeling approach.
                
                The proposed USNRT has a computational complexity proportional to the tree depth, and pruning is unnecessary. As we can see in the experimental section, generally, a tree with depth 2, 3, or at most 4 is enough to produce good performance. Thus, USNRT's computational cost is satisfactory. We apply it on 17 extensive UCI datasets of various sizes and compare it with existing representative models. The experimental results show that USNRT achieves superior performance over others. Furthermore, comprehensive visualization and analysis indicate that such a tree-structured learning method is successful and interpretable, and has merits as stated.

We summarize our main contributions as follows.
\begin{enumerate}
	\item We propose a novel method for tree-structured learning of uncertainty. The novelty lies in the new splitting rule of the tree designed using a statistical test on the residuals, which has shown its effectiveness and benefits in the paper.
	\item Although many recent studies focus on epistemic uncertainty estimation, making studying it extremely popular, we show that aleatory uncertainty estimation (of regression task) is far from being well resolved. We make a further step in variance network methodology, showing that the performance can be substantially improved by the proposed USNRT.
	\item Our method can be easily extended to an ensemble version (change the network initialization seed), so can estimate the total uncertainty including the epistemic. Compared to recent popular and similar methods, the proposed USNRT Ensemble shows promising performance on total uncertainty estimation.
	 (However, there are some unsolved debates on epistemic uncertainty estimation, e.g., in \cite{lahlou2021deup}, hence it is not our main focus.)
	\item Our approach is intuitive and well-motivated. Through comprehensive visualization and analysis, we find that uncertainty heterogeneity indeed exists in many datasets and  we demonstrate that USNRT can learn the uncertainty heterogeneity. Its performance is closely related to an uncertainty heterogeneity measure defined by us.
	These results may have many implications for future research and industrial applications. %This empirical finding is absolutely new in recent machine learning literature.
	\item The proposed method is robust to hyper-parameters, is easy to interpret, and has acceptable computational cost.
	%\item We have released the code and data in \url{https://github.com/xingyan-fml/usnrt}. Everyone is welcome to check and run our code to verify the substantial improvement of aleatory uncertainty estimation.
\end{enumerate}
        Code and data for reproducing all the experimental results are available in the link\footnote{\href{https://github.com/xingyan-fml/usnrt}{https://github.com/xingyan-fml/usnrt}}. They are for tabular data currently.

        \section{Related Works}
        
        We review the related works of both variance network-based uncertainty quantification and tree-based space partitioning. Our method is closely related to both.
        
        \subsection{Aleatory Uncertainty and Epistemic Uncertainty}
         
        The source of uncertainty can be generally categorized into two kinds, epistemic uncertainty and aleatory uncertainty \cite{der2009aleatory}. Epistemic uncertainty or systematic uncertainty is caused by a lack of knowledge, and can be reduced by increasing additional information \cite{hullermeier2021aleatoric, brando2019modelling}. For example, one can add prior information to the model through the Bayesian approach \cite{blundell2015weight, graves2011practical, hernandez2015probabilistic, li2015stochastic, maddox2019simple, gal2016dropout, gal2017concrete, kingma2015variational}, or expanding the training data.
        
        Aleatory uncertainty or statistical uncertainty refers to randomness which cannot be reduced by any additional information. This type of uncertainty can be quantified by outputting the full conditional distribution, rather than the point-wise prediction. Many popular methods are proposed to estimate the aleatory uncertainty, such as variance networks \cite{lakshminarayanan2017simple, skafte2019reliable, zhao2020individual, cui2020calibrated}, quantile regressions \cite{chung2021beyond, feldman2021improving, romano2019conformalized}, conformal predictions \cite{shafer2008tutorial, papadopoulos2011regression, romano2019conformalized}, and distribution predictions \cite{zhou2021estimating}. 
        %In our regression problem, data heterogeneity divides the data into different regions, in which the models and aleatory uncertainties are different. Our method applies deep learning to estimate the models of different regions and quantify the corresponding aleatory uncertainties.
        
        Generally in the variance methodology, Bayesian and ensemble methods estimate the total uncertainty of the two, and non-Bayesian methods that output only one mean and one variance estimate aleatory uncertainty only. We will avoid mentioning this many times in the rest of the paper.
        In this paper, our method can originally estimate aleatory uncertainty, and an ensemble version can estimate the total uncertainty.
        
        \subsection{Variance Networks and Uncertainty Quantification}
        
        \textbf{HNN and Deep Ensemble.} With the Gaussian assumption, Heteroscedastic Neural Network (HNN) tries to output its mean and variance. However, the training is not straightforward.
        There were some works proposing various training schemes. For instance, \cite{lakshminarayanan2017simple} used one network to output both the mean and the variance, and adopted the negative log-likelihood loss and a gradient clipping strategy.
        \cite{zhao2020individual} combined the negative log-likelihood loss with a new mPAIC loss and trained randomized regression functions to achieve individual calibration. \cite{skafte2019reliable} proposed a new mini-batching scheme and trained a mean network and a variance network separately and alternately. Furthermore, multiple means and variances given by several HNNs with different random initializations can be aggregated to form a mixture of Gaussian distributions, which is called the Deep Ensemble method \cite{lakshminarayanan2017simple}.
        
        \textbf{Dropout-based Bayesian methods.} The Bayesian approach has tried early to estimate the posterior predictive variance. Bayesian methods first give the network's parameters a prior distribution, and then given the training data, they compute the posterior distribution of the parameters and the label. But they have a prohibitive computational cost. Various Bayesian approximations were developed to overcome the computational issue. \cite{hernandez2015probabilistic} made use of the expectation propagation to estimate the posterior distribution by a set of local approximations. MC Dropout \cite{gal2016dropout} and Concrete Dropout \cite{gal2017concrete} used dropout as a Bayesian approximation, which is much easier to implement.% The accuracy of these methods relies on the degree of approximation and the exactness of the prior distribution. 
        
        \textbf{Deep Evidential Regression.} \cite{amini2020deep} proposed a non-Bayesian method to estimate a continuous label's evidence in order to learn both aleatoric and epistemic uncertainty. They placed evidential priors over the original Gaussian likelihood and trained the network to predict the hyperparameters of the evidential distribution. They additionally imposed priors as the model regularization for letting the predicted evidence aligned with the correct output. Overall, the method allows efficient and scalable uncertainty learning.
        
        \textbf{Tree models with variance estimation.} On structured data, it is common to use ensemble tree models to achieve good prediction accuracy. They can also estimate a variance as the uncertainty measurement. Random Forest \cite{breiman2001random} is a famous ensemble tree model and can return the standard deviation accompanied with the prediction value. Extremely Randomized Trees (ExTra) \cite{geurts2006extremely} 
        is another popular one which randomly selects the split points and then chooses the best split. ExTra can return the standard deviation (possibly larger than that given by Random Forest) with the prediction value too. There are some other differences between Random Forest and ExTra. We will use both of them to obtain the mean and variance for quantifying aleatory uncertainty.
        
        \subsection{Space Partitioning and Tree Models}
        
        Space partitioning is inherent in tree models, where linear and hard (versus soft) partitioning is adopted. Many tree-based models have been proposed, most of which recursively partition the feature space in an axis-parallel way. Most notably, CART \cite{breiman2017classification} constructs the tree structure by greedily partitioning the data to decrease the sum of squared errors and uses constant-valued models within nodes. Afterwards, some studies contribute to upgrading the constant-valued models. For example, \cite{chaudhuri1994piecewise} proposed piecewise polynomial regression tree (SUPPORT) which recursively utilizes the estimated residuals’ signs to divide data into two parts. It determines the best split by conducting a statistical test. Finally, a polynomial model is fitted within each leaf node. Segmented linear regression tree (SLRT) \cite{zheng2019partitioning} fits a linear model at each node and calculate the cumulative Kendall's $\tau$ rank correlation between covariates and estimated residuals to select the best partition. % They use a heuristic splitting criterion to ensure computational efficiency as well. However, the main weakness of this study is its failure to capture the nonlinearity and interaction of data because they use linear models at leaf nodes. 
        Piecewise polynomial or linear tree models are useful, but may not be attractive given the complexity, nonlinearity, and heterogeneity of large real-world datasets. 
        
        When deep learning is extremely popular and has excellent expressive power, as far as we know, there were few works that combine the tree models with neural networks such that node-specific networks are employed for leaf nodes. Existing works that try to combine the two worlds include \cite{kontschieder2015deep,humbird2018deep,yang2018deep,zhu2018learning,ke2019deepgbm,wan2020nbdt}. Most of them focused on specific tasks such as recommender systems \cite{zhu2018learning}, online predictions \cite{ke2019deepgbm}, and network initialization \cite{humbird2018deep}.
        In \cite{kontschieder2015deep,wan2020nbdt}, they combined the final linear layer of a neural network with a differentiable decision tree model, which is quite different with the angle of data heterogeneity  and space partitioning in this paper. 
        
        \subsection{Other Local Models and Tree Models}
        
        There were similar ideas of partitioning the space proposed, such as in \cite{gramacy2008bayesian,lee2023partitioned}. The treed Gaussian Process models in  \cite{gramacy2008bayesian} used decision trees for
partitioning as well. Independent local Gaussian Process was fitted on the subregion of each leaf node. \cite{lee2023partitioned} proposed a partitioned active learning method for data acquisition and modeling of heterogeneous systems, which partitions the design space based on heterogeneous features and seeks the most uncertain subregions.
        
        Besides, our work shares some similarities with Hierarchical Mixture of Experts (HMEs) \cite{jordan1994hierarchical} in a small extent. HMEs are soft local-regression models which adopt multiple local models and combine them in a probabilistic way, like Mixture of Gaussians for clustering. Generally, EM algorithm needs to be used to learn the parameters of such probabilistic models. This may be computationally inefficient when neural networks are used as local models. While we use the hard-splitting tree structure, it will be more natural to incorporate neural networks.
        
        Tree models are a big family. Chi-Square Automatic Interaction Detector (CHAID) proposed by \cite{kass1980exploratory} determines the best multiway splits by conducting statistical tests too. However, CHAID is preferred when there are many categorical variables and is more suitable for target selection problems. The representatives of other ensemble tree models include Gradient Boosting Decision Tree (GBDT) \cite{friedman2001greedy} and LightGBM \cite{ke2017lightgbm}. They are both boosting-type and generally do not output a variance for the prediction.
        
        \subsection{Other Types of Methods for Uncertainty Quantification}
        
        Except for variance-based methods for uncertainty quantification in regression, there are still other types of methods such as quantile-based interval constructions and new loss functions designed for predicting the whole distribution.
        
        Estimating quantiles does not need any parametric assumptions on the distribution. Two quantiles of symmetric levels (about probability 0.5) can form the prediction interval we need. Quantile regression \cite{koenker1978regression,koenker2001quantile} has a long history in statistics and econometrics, and pinball loss is the core of the literature. Recently in machine learning, \cite{tagasovska2019single} proposed Simultaneous Quantile Regression (SQR) which models the quantile function (inverse of CDF) as a partial neural network and estimates all the conditional quantiles of the label simultaneously. \cite{feldman2021improving} proposed orthogonal quantile regression which utilizes the independence between the interval length and the indicator of miscoverage to design a new regularization term. 
        \cite{chung2021beyond} proposed several new loss functions or training procedures to overcome the shortcomings of pinball loss and obtain intervals with better calibration and sharpness.
        \cite{romano2019conformalized} proposed a novel conformalized quantile regression with the advantages of both conformal prediction and quantile regression.%, which aims to construct prediction intervals that can achieve finite sample calibration. %Like other quantile methods, it estimates a pair of quantiles each time, and cannot approximate a full distribution like the methods mentioned before do.
        
        Besides, some new loss functions have been proposed to estimate the whole conditional distribution for quantifying uncertainty. \cite{cui2020calibrated} used Maximum Mean Discrepancy as a distribution matching strategy to construct calibrated regression. \cite{zhou2021estimating} proposed Collaborating Networks which use two distinct networks to represent the quantile function and its inverse (CDF), and proposed novel loss functions to learn them.
         \cite{brando2019modelling} estimated heterogeneous conditional distributions by combining uncountable asymmetric Laplacians. %Some studies measure uncertainty by conformal predictions \cite{shafer2008tutorial, papadopoulos2011regression}.

        \section{Theoretical Justification}
        
        We first justify the usefulness of space partitioning theoretically, by showing that if the data-generating process is heterogeneous, partitioning the feature space always yields a smaller expected loss or learning objective for quantifying uncertainty.
        
        \begin{proposition}\label{thm:least.square}
        Suppose two uncountable infinite sets $\mathcal{D}_0\bigcup\mathcal{D}_1=\mathbb{R}^d$, $\mathcal{D}_0\bigcap\mathcal{D}_1=\emptyset$, and the closures $\widebar{\mathcal{D}_0}\bigcap\widebar{\mathcal{D}_1}=\{x\in\mathbb{R}^d|B(x)=0\}\ne\emptyset$, where $B(\cdot)\in C^0(\mathbb{R}^d)$. Assume $\mathbf{y}_0=f_0\left(\mathbf{X}_0\right)+\boldsymbol{\varepsilon}_0$, $\mathbf{y}_1=f_1\left(\mathbf{X}_1\right)+\boldsymbol{\varepsilon}_1$, where $f_0\in C^0(\widebar{\mathcal{D}_0})$, $f_1\in C^0(\widebar{\mathcal{D}_1})$, and $\mathbf{X}_0,\mathbf{X}_1,\boldsymbol{\varepsilon}_0,\boldsymbol{\varepsilon}_1$ are mutually independent random variables: $\mathbf{X}_0: \Omega\to\mathcal{D}_0$, $\mathbf{X}_1: \Omega\to\mathcal{D}_1$, and $\boldsymbol{\varepsilon}_0, \boldsymbol{\varepsilon}_1$ are noise variables with constant parameters.
        If another independent $Z\sim\text{Bernoulli}(p)$, let $\mathbf{X} =Z\mathbf{X}_1+\left(1-Z\right)\mathbf{X}_0$ and $\mathbf{y} =Z\mathbf{y}_1+\left(1-Z\right)\mathbf{y}_0$, given a proper loss function $l$ (such as least square), we have
        \begin{align}\label{eqn:thm:least.square}
                & (1-p)\min_{f\in\mathcal{F}}\mathbb{E}\left[l\left(\mathbf{y}_0,f(\mathbf{X}_0)\right)\right]+ p\min_{f\in\mathcal{F}}\mathbb{E}\left[l\left(\mathbf{y}_1,f(\mathbf{X}_1) \right)\right] \nonumber \\
                & \le\min_{f\in\mathcal{F}}\mathbb{E}\left[l\left(\mathbf{y},f\left(\mathbf{X}\right)\right)\right],
        \end{align}
        where $\mathcal{F}$ is a function set (such as neural networks) satisfying $\mathcal{F}\subset C^0(\mathbb{R}^d)$.
        \end{proposition}
        
        \noindent
        {\it Proof:} According to the double expectation formula, we have
        \begin{align*}
                & \mathbb{E}\left[l\left(\mathbf{y},f\left(\mathbf{X}\right)\right)\right] \\
                & =\mathbb{E}\left[l\left(Z\mathbf{y}_1+\left(1-Z\right)\mathbf{y}_0,f\left(Z\mathbf{X}_1+\left(1-Z\right)\mathbf{X}_0\right)\right)\right] \\
                & = \mathbb{E}\left[\mathbb{E}\left[l\left(Z\mathbf{y}_1+\left(1-Z\right)\mathbf{y}_0,f\left(Z\mathbf{X}_1+\left(1-Z\right)\mathbf{X}_0\right)\right)|Z\right]\right] \\
                & = \left(1-p\right)\mathbb{E}\left[l\left(\mathbf{y}_0,f\left(\mathbf{X}_0\right)\right)\right] + p\mathbb{E}\left[l\left(\mathbf{y}_1,f\left(\mathbf{X}_1\right) \right)\right] \\
                & \ge (1-p)\min_{f\in\mathcal{F}}\mathbb{E}\left[l\left(\mathbf{y}_0,f\left(\mathbf{X}_0\right)\right)\right]+ p\min_{f\in\mathcal{F}}\mathbb{E}\left[l\left(\mathbf{y}_1,f\left(\mathbf{X}_1\right) \right)\right].
        \end{align*}
        Therefore, if we take the minimum of $\mathbb{E}\left[l\left(\mathbf{y},f\left(\mathbf{X}\right)\right)\right]$ for $f\in\mathcal{F}$, we can finish the proof.\hfill$\blacksquare$
        %\begin{equation*}
        %	\min_{f\in\mathcal{F}}\mathbb{E}\left[l\left(\mathbf{y},f\left(\mathbf{X}\right)\right)\right]\ge(1-p)\min_{f_0\in\mathcal{F}}\mathbb{E}\left[l\left(\mathbf{y}_0,f\left(\mathbf{X}_0\right)\right)\right]+ p\min_{f_1\in\mathcal{F}}\mathbb{E}\left[l\left(\mathbf{y}_1,f\left(\mathbf{X}_1\right) \right)\right]
        %\end{equation*}\hfill$\blacksquare$

        \begin{corollary}\label{thm:likelihood}
        Under the conditions of Proposition \ref{thm:least.square}, we consider the heteroscedasticity situation. Now we assume $\mathbf{y}_0 = f_0\left(\mathbf{X}_0\right)+\sigma_0\left(\mathbf{X}_0\right)\boldsymbol{\varepsilon}_0$ and $\mathbf{y}_1=f_1\left(\mathbf{X}_1\right)+\sigma_1\left(\mathbf{X}_1\right)\boldsymbol{\varepsilon}_1$, where $\sigma_0\in C^0(\widebar{\mathcal{D}_0})$, $\sigma_1\in C^0(\widebar{\mathcal{D}_1})$, given a proper loss function $l(\cdot,\cdot,\cdot)$ between $y$ and $(f(x),\sigma(x))$ (such as negative log-likelihood), we have
        \begin{align}\label{eqn:thm:likelihood}
                & (1-p)\min_{f\in\mathcal{F},\sigma\in\mathcal{F}'}\mathbb{E}\left[l\left(\mathbf{y}_0,f(\mathbf{X}_0),\sigma(\mathbf{X}_0)\right)\right] \nonumber \\
                & + p\min_{f\in\mathcal{F},\sigma\in\mathcal{F}'}\mathbb{E}\left[l\left(\mathbf{y}_1,f(\mathbf{X}_1),\sigma(\mathbf{X}_1) \right)\right] \nonumber \\
                & \le\min_{f\in\mathcal{F},\sigma\in\mathcal{F}'}\mathbb{E}\left[l\left(\mathbf{y},f\left(\mathbf{X}\right),\sigma(\mathbf{X})\right)\right],
        \end{align}
        where $\mathcal{F}$ and $\mathcal{F}'$ are function sets satisfying $\mathcal{F}\subset C^0(\mathbb{R}^d)$, $\mathcal{F}'\subset C^0(\mathbb{R}^d)$.\hfill$\blacksquare$
        \end{corollary}

        \begin{proposition}\label{thm:inequality.strict}
        Under the conditions of Corollary \ref{thm:likelihood}, the inequality in Corollary \ref{thm:likelihood} is strict if we further assume the following conditions are satisfied:
        \begin{enumerate}
        \item $\mathbf{X}_0$ and $\mathbf{X}_1$ have positive densities over $\mathcal{D}_0$ and $\mathcal{D}_1$ respectively;
        \item either $f_0,f_1$ or $\sigma_0,\sigma_1$ are discontinuous at some points on the boundary between $\mathcal{D}_0$ and $\mathcal{D}_1$;
        \item $\mathcal{F}$ is closed under addition, and the functions composed by $Wx+b$ and $\max(x,0)$ with at most $T_{max}$ times and $d_{max}$ internal dimensions, are in $\mathcal{F}$. $f\in\mathcal{F}'$ if $f\in\mathcal{F}$ and $f>0$. $T\circ f\in\mathcal{F}'$ if $f\in\mathcal{F}$ and $T\in C^0(\mathbb{R})$, $T>0$.
        %\item there exists a sequence of functions $f_0^{(k)}\in\mathcal{F}$ such that $\mathbb{E}[|f_0^{(k)}(\mathbf{X}_0) - f_0(\mathbf{X}_0)|] \to 0$, $k\to\infty$, and similar sequences $f_1^{(k)}$, $\sigma_0^{(k)}$, $\sigma_1^{(k)}$ exist for $f_1$, $\sigma_0$, $\sigma_1$.
        \end{enumerate}
        The similar conclusion applies on the inequality in Proposition \ref{thm:least.square} too.
        Moreover, the conclusion applies for other activation functions beyond $\max(x,0)$, such as Sigmoid and Tanh.
        \end{proposition}
        
        \noindent
        {\it Proof Sketch:} Without loss of generality, we assume $f_0$ and $f_1$ are discontinuous at the point $x^*$ on the boundary between $\mathcal{D}_0$ and $\mathcal{D}_1$. Suppose $\min_{f\in\mathcal{F},\sigma\in\mathcal{F}'}\mathbb{E}\left[l\left(\mathbf{y},f\left(\mathbf{X}\right),\sigma(\mathbf{X})\right)\right]$ is solved at $f^*, \sigma^*$, because $f^*$ is a continuous function and $f_0,f_1$ are continuous on $\widebar{\mathcal{D}_0}$ and $\widebar{\mathcal{D}_1}$ respectively, we can find a hypercube $H$ centered at $x^*$ with length $\delta$ such that:
        \begin{align*}
                &  | f^*(x) -f^*(x^*) | <  \epsilon/8, \quad  x\in H, \\
                &  | f_0(x) -f_0(x^*) | <  \epsilon/8, \quad  x\in H\bigcap\mathcal{D}_0, \\
                &  | f_1(x) -f_1(x^*) | <  \epsilon/8, \quad  x\in H\bigcap\mathcal{D}_1,
        \end{align*}
        where $\epsilon=|f_1(x^*)-f_0(x^*)|$. This can easily yield $| f^*(x) -f_0(x) | >  \epsilon/4$ for all $x\in H\bigcap\mathcal{D}_0$, or, $| f^*(x) -f_1(x) | >  \epsilon/4$ for all $x\in H\bigcap\mathcal{D}_1$. Without loss of generality, we assume the former is true, then we can further find an $f_2\in\mathcal{F}$ according to the condition 3), such that $f_2(x)=0$ when $x\notin H$, and $f^*+f_2$ gives: 
        \begin{align*}
        & \mathbb{E}\left[l\left(\mathbf{y}_0,(f^*+f_2)(\mathbf{X}_0),\sigma^*(\mathbf{X}_0)\right)\right] \\
        < & \mathbb{E}\left[l\left(\mathbf{y}_0,f^*(\mathbf{X}_0),\sigma^*(\mathbf{X}_0)\right)\right].
        \end{align*}
        This inequality holds because $\mathbf{X}_0$ and $\mathbf{X}_1$ have positive densities around $x^*$ and $l$ is a proper loss such that a better $f\in\mathcal{F}$ will lead to a smaller expected risk. 
        With this inequality,  we can finish the proof. The full proof is in Appendix \ref{proof:inequality.strict}.\hfill$\blacksquare$
        
        \subsection{Explanations on Theorems}\label{explanations}

We explain all the theorems in the following:
\begin{enumerate}
	\item Our basic assumption is that the data is generated by two different equations in two feature regions, with a probability $p$ to mix them (see the description in Proposition \ref{thm:least.square}). In this situation, we have proven that learning separately in the two regions will yield a smaller (or equal) expected risk compared to learning for the whole, see Equation \eqref{eqn:thm:least.square} in Proposition \ref{thm:least.square}.
	\item Now back to the uncertainty quantification problem. Heteroscedastic Neural Network assumes conditional Gaussian for the data and uses negative log-likelihood as the loss function to learn the conditional mean and variance. We thereby extend Proposition \ref{thm:least.square} to the case of HNN in which two models $f$ and $\sigma^2$ predict conditional mean and conditional variance respectively, see Equation \eqref{eqn:thm:likelihood} in Corollary \ref{thm:likelihood}. This extension is straightforward. Now the expected risk in Proposition \ref{thm:least.square} becomes the expected negative log-likelihood of HNN.
	\item At last, we want to show that if the data-generating equations in the two regions are discontinuous at the boundary, i.e., $f_0$ and $f_1$ (or $\sigma_0$ and $\sigma_1$) are discontinuous at the boundary, we can definitely obtain a strictly smaller expected loss (expected negative log-likelihood) when  learning separately in the two regions, see Proposition \ref{thm:inequality.strict}.
	\item The theorems are consistent with the intuition that neural networks (which are continuous functions) cannot do better than tree models (which are discontinuous functions) on tabular data because tabular data usually presents discontinuous predictive patterns.
	\item Discontinuity is only one sufficient condition for our model to learn better (inequalities in Equation \eqref{eqn:thm:least.square} and \eqref{eqn:thm:likelihood} become strict). 
	Another sufficient condition is that $(f_0,\sigma_0)$ and $(f_1,\sigma_1)$ in the two regions are much different (heterogeneous), hence they cannot be learned well as a whole with a relatively restricted function class. Despite of the Universal Approximation Theorem which states the power of infinitely large neural network, it is infeasible to build an extremely large network to learn across all regions. Under this circumstance, learning separately in two regions will be less difficult obviously. %The detailed conditions in mathematical forms are worthy of being explored in the future.
\end{enumerate}
        We name both the two situations, i.e., the discontinuity and the difference between data-generating equations, as data or uncertainty heterogeneity. We will show in the experiments the empirical evidence supporting the statements here.
        
                \section{Methodology}
                
                In this section, we describe how to construct the proposed Uncertainty-Splitting Neural Regression Tree (USNRT). The algorithm details are also presented in Algorithm \ref{DT}.% When finishing the tree structure, the prediction and uncertainty estimation in the leaf nodes are left in the next section.
                
                \subsection{Framework}
                
                We focus on the regular regression problem between the outcome $\mathbf{y}$ and the features $\mathbf{X} = (\mathbf{X}_1, \dots, \mathbf{X}_d)^{\top}$, given the dataset $\left\{\left(\mathbf{X}\left(i\right), \mathbf{y}\left(i\right)\right)\right\}_{i=1}^{N}$, where $\mathbf{X}\left(i\right)\in\mathcal{X}$ and $\mathcal{X}$ is the feature space.
                We hope to build a model that captures the heterogeneity across $M$ disjoint regions $\left\{\mathcal{R}_m\right\}_{m=1}^{M}$ partitioned from $\mathcal{X}$, satisfying $\mathcal{X}=\bigcup_{m=1}^{M}\mathcal{R}_m$. For different regions, the prediction models and the uncertainty estimation models are distinct. % But they keep homogeneous when $\mathbf{X}$ varies within one fixed region $\mathcal{R}_m$. 
                To be formal, our model is
                \begin{equation}
                        \mathbf{y}=\sum_{m=1}^{M}\left(f_m(\mathbf{X})+ \sigma_m(\mathbf{X}) \boldsymbol{\varepsilon}_m\right) \mathbb{I}_{\left\{\mathbf{X} \in \mathcal{R}_{m}\right\}},
                \end{equation}
                where $f_m(\mathbf{X})$ predicts $\mathbf{y}$ given $\mathbf{X}$ locating in the region $\mathcal{R}_{m}$, and $\sigma_m(\mathbf{X})$ denotes the individual uncertainty level. We assume $\boldsymbol{\varepsilon}_m$ is standard Gaussian-distributed.
                
                We further assume that the regions $\left\{\mathcal{R}_m\right\}_{m=1}^{M}$ are obtained by recursively partitioning $\mathcal{X}$ in an axis-parallel way. In the context of tree models, that is, at each internal node $t$, we choose both the best split variable $\mathbf{X}_{\hat{k}}$ and the best split value $\hat{a}$ under some splitting criteria to obtain two child nodes: $\mathbf{X}_{\hat{k}} \le \hat{a}$ and $\mathbf{X}_{\hat{k}} > \hat{a}$. 
                
                \subsection{Recrusive Partitioning with Uncertainty Heterogeneity}
                
                 When constructing the tree with training data, at each internal node $t$, we first train a neural network model $\hat{f}_{\text{split}}(\mathbf{X})$ with the data in the node using the mean squared error loss. %The parameters are optimized to minimize the mean squared error (MSE):
        %	 \begin{equation}
        %	 	\frac{1}{|\text{train}_t|}\sum_{i\in \text{train}_t}\left(y\left(i\right) - \hat{f}_{\text{split}}\left(\mathbf{X}\left(i\right)\right)\right)^2,
        %	 \end{equation}
        %	 where $\text{train}_t$ is the training set of node $t$. 
                 Then we can obtain the residual $\hat{\mathbf{e}} = \mathbf{y} - \hat{f}_{\text{split}}(\mathbf{X})$. 
                 
                 If the region of node $t$ is partitioned into two sub-regions, we want to check whether the overall uncertainty levels in this two sub-regions are heterogeneous. If yes, the node $t$ will be divided to reduce the difficulty of modeling. To measure the overall uncertainty level, we use the sample variance of the residual $\hat{\mathbf{e}}$. Therefore, our goal is simplified to verifying whether we can partition the residuals $\{\hat{\mathbf{e}}(i)\}$ into two groups, such that the residuals' variances in the two groups are significantly different. We can conduct such an examination with the help of Levene's test \cite{levene1961robust}, which is a statistical tool to test the equality of variances of two groups of observations.
                
                For an arbitrary split choice with split variable $\mathbf{X}_k$ and split value $a$, there will be two separate index sets of the data samples: $\mathcal{I}_{L} = \left\{i | \mathbf{X}_k\left(i\right) \leq a\right\}$ and  $\mathcal{I}_{R} = \left\{i | \mathbf{X}_k\left(i\right) > a\right\}$. Correspondingly, we can obtain two groups of residuals: $\hat{\mathbf{e}}_{L} = \left\{\hat{\mathbf{e}}\left(i\right)|i \in \mathcal{I}_{L}\right\}$ and $\hat{\mathbf{e}}_{R} = \left\{\hat{\mathbf{e}}\left(i\right)|i \in \mathcal{I}_{R}\right\}$. Let $\bar{e}_{*}$ and $\hat{\sigma}^2_{*}$ denote the sample mean and sample variance of $\hat{\mathbf{e}}_{*}$, where $*$ represents the subscript $L$ or $R$. In Levene's test \cite{levene1961robust}, the null hypothesis is the equality of the two variances:
                \begin{equation}
                        H_0:\quad{\sigma}_{L}^2 = {\sigma}_{R}^2.
                \end{equation}
                We define $\mathbf{z}_{*} = \left\{ | \hat{\mathbf{e}}\left(i\right) - \bar{e}_{*}| \big| i \in \mathcal{I}_{*} \right\} $, and let $\bar{z}_{*}$ and $w_{*}^2$ denote the sample mean and sample variance of $\mathbf{z}_{*}$. Then the pooled variance is defined as $w_\text{pool}^2 = \frac{(|\mathcal{I}_{L}| - 1)w_{L}^2 + (|\mathcal{I}_{R}| - 1)w_{R}^2}{|\mathcal{I}_{L}| + |\mathcal{I}_{R}| - 2}$. Following \cite{chaudhuri1994piecewise}, the Levene's test statistic is defined as follows:
                \begin{equation}
                        T_\text{Levene} = \frac{\bar{z}_{L} - \bar{z}_{R}}{w_\text{pool}\sqrt{|\mathcal{I}_{L}|^{-1} + |\mathcal{I}_{R}|^{-1}}}.
                \end{equation}
                
                This test statistic follows  a Student's $t$-distribution with $|\mathcal{I}_{L}| + |\mathcal{I}_{R}| - 2$ degrees of freedom if the null hypothesis is supported. We can compute the corresponding $p$-value, denoted as $p(k, a)$, for each split choice $\mathbf{X}_k$ and $a$. Among all these $p$-values obtained, we choose the variable index $\hat{k}$ and the split value $\hat{a}$ corresponding to the smallest one as our best split choice:
                \begin{equation}
                        p_{\text{best}} = p(\hat{k}, \hat{a}) = \min_{k, a}p(k, a).
                \end{equation}
                If $p_{\text{best}}$ is less than a given significance level $\alpha$, we can infer a significant difference between the two groups of residuals. Then, we partition the region into two sub-regions: $\mathbf{X}_{\hat{k}} \leq \hat{a}$ and $\mathbf{X}_{\hat{k}} > \hat{a}$, in which the residuals' variances are the most significantly different. So, this node is divided and the data samples in the node are divided into two parts correspondingly. Recursively doing this, we can build a tree.
                
                Our splitting criteria is a little like that of SUPPORT \cite{chaudhuri1994piecewise} because both use statistical tests. However, SUPPORT only uses the signs of estimated residuals which are much less informative. Instead, our method utilizes the estimated residuals in a more data-driven way and is more intuitive.
                
                \subsection{Stopping Rules}
                
                We set two stopping rules to determine whether to divide the node, or to stop and return the current node as a leaf node. First, we ensure that in every node there must be at least $N_{\min}$ samples in the tree building. Hence, in the split selection, the split value $a$ should satisfy some constraints to ensure this. Moreover, it helps stop the tree growth. When a node has a sample size less than $2N_{\min}$, it will not be divided anymore. The second stopping rule is that $p_{\text{best}}$ obtained from Levene's test should be no more than the given significance level $\alpha$, otherwise, the current node will not be divided. 
                
                The obtained tree can benefit from these two stopping rules. Most importantly, unlike in traditional tree models, pruning is unnecessary for USNRT. In traditional tree models like CART, the splitting criteria is minimizing the sum of square errors in the two child nodes. The deeper the tree is, the more leaf nodes there are, and the more accurate the performance on  training data is. Thus, there is a need to balance the accuracy and the complexity of the tree to overcome over-fitting and achieve good generalization. However, our splitting goal is to discover heterogeneity across regions and reduce the difficulty of modeling the data in child nodes. At every leaf node, neural networks will be trained to fit the data well in that node only. Actually, the stopping rules let the tree grow adaptively without the risk of being too shallow or too deep, making the pruning unnecessary. As we will see in the experiments, $10$ leaf nodes for USNRT are enough for producing good performance. A consequent benefit is that USNRT will not be computationally costly. The summary of the tree construction is in Algorithm \ref{DT}.
                
                \begin{algorithm}[tb]
                        \caption{Uncertainty-Splitting Neural Regression Tree Construction}
                        \label{DT}
                        \begin{algorithmic}[1]
                                \Require Dataset $\left\{\left(\mathbf{X}\left(i\right),\mathbf{y}\left(i\right)\right)\right\}_{i=1}^N$, significance level $\alpha$, minimum sample size of leaf nodes $N_{\text{min}}$, and hyper-parameters of the splitting neural network.
                                %\Ensure The final model $\hat{f}(\mathbf{X}) = \sum_{m=1}^{M}\hat{f}_m(\mathbf{X})\mathbf{1}_{\left\{\mathbf{X} \in \mathcal{R}_{m}\right\}}$, where $\mathcal{R}_{m}$ is the region of the $m$-th leaf node, having a prediction variance $\hat{\sigma}_m^2$.
                                \Ensure A grown tree which partitions the feature space to $\mathcal{X}=\bigcup_{m=1}^{M}\mathcal{R}_m$, where $\mathcal{R}_{m}$ is the region of the $m$-th leaf node.
                                \State Initialize: $m = 0$.
                                \State Denote the current node as $t$, and the index set of all samples at this node as $\mathcal{I}_{t}$. %root region $\mathcal{R}_t = \left\{\mathbf{X}\left(i\right) | i \in \mathcal{I}_{t}\right\}$
                                \If{$|\mathcal{I}_{t}| < 2N_{\min} $}
                                \State $m = m+1$, denote the region of this node as  $\mathcal{R}_{m}$, and return as a leaf node.
                                %\State Fit a leaf neural network $\hat{f}_{m}(\mathbf{X})$ using samples in node $t$. $\hat{\sigma}_m^2 = $ the residuals' variance.
                                %\State $\hat{f}(\mathbf{X}) = \hat{f}(\mathbf{X}) + \hat{f}_{m}(\mathbf{X})\mathbf{1}_{\left\{\mathbf{X} \in \mathcal{R}_{m}\right\}}$.
                                \Else
                                \State Fit a splitting neural network $\hat{f}_{\text{split}}(\mathbf{X})$ with MSE loss using samples in node $t$, and get the residuals $\hat{\mathbf{e}}\left(i\right) = \mathbf{y}\left(i\right) - \hat{f}_{\text{split}}(\mathbf{X}\left(i\right)), i\in\mathcal{I}_t$.
                                \For{$k = 1$ to $d$}
                                \For{$a$ in $\text{sorted}\left\{\mathbf{X}_k\left(i\right)|i\in \mathcal{I}_t\right\}$}
                                \State Divide the index set into $\mathcal{I}_{L} = \left\{i\in \mathcal{I}_t | \mathbf{X}_k\left(i\right) \leq a\right\}$ and $\mathcal{I}_{R} = \left\{i\in \mathcal{I}_t | \mathbf{X}_k\left(i\right) > a\right\}$.
                                \State \textbf{if} $|\mathcal{I}_{L}|<N_{\text{min}} $ or  $|\mathcal{I}_{R}|<N_{\text{min}} $ \textbf{then} continue
                                \State Divide the residuals into $\hat{\mathbf{e}}_{L} = \left\{\hat{\mathbf{e}}\left(i\right)|i \in \mathcal{I}_{L}\right\}$ and $\hat{\mathbf{e}}_{R} = \left\{\hat{\mathbf{e}}\left(i\right)|i \in \mathcal{I}_{R}\right\}$.
                                \State Do the Levene's test for the variances of $\hat{\mathbf{e}}_{L}$ and $\hat{\mathbf{e}}_{R}$ and compute the $p$-value, $p(k, a)$. 
                                \EndFor
                                \EndFor
                                \State Get $p_{\text{best}} = p(\hat{k}, \hat{a}) = \min_{k, a}p(k, a)$.
                                \If{$p_{\text{best}} > \alpha$}
                                \State $m = m+1$, denote the region of this node as  $\mathcal{R}_{m}$, and return as a leaf node.
                                %\State Fit a leaf neural network $\hat{f}_{m}(\mathbf{X})$ using samples in node $t$. $\hat{\sigma}_m^2 = $ the residuals' variance.
                                %\State $\hat{f}(\mathbf{X}) = \hat{f}(\mathbf{X}) + \hat{f}_{m}(\mathbf{X})\mathbf{1}_{\left\{\mathbf{X} \in \mathcal{R}_{m}\right\}}$.
                                \Else
                                \State Split the node $t$ into two child nodes $t_{L}$ and $t_{R}$, where $\mathcal{I}_{t_L} = \left\{i\in\mathcal{I}_t | \mathbf{X}_{\hat{k}}\left(i\right) \leq \hat{a}\right\} $ and $\mathcal{I}_{t_R} = \left\{i\in\mathcal{I}_t | \mathbf{X}_{\hat{k}}\left(i\right) > \hat{a}\right\}$.
                                \State  Let $t = t_{L}$ and repeat Step 2 -- 23.
                                \State  Let $t = t_{R}$ and repeat Step 2 -- 23.
                                \EndIf
                                \EndIf
                        \end{algorithmic}
                \end{algorithm}
        
                \subsection{Prediction and Uncertainty Estimation}
                
                The prediction and uncertainty estimation are made in every leaf region $\mathcal{R}_{m}$ separately. When $\mathbf{X}\in\mathcal{R}_{m}$, two neural networks $\hat{f}_m(\mathbf{X})$ and $\hat{\sigma}_m(\mathbf{X})$ are trained using the data in $\mathcal{R}_{m}$ only, to give the prediction and uncertainty estimation of $\mathbf{y}$. %Within the region of each leaf node, we will employ a neural network model to make predictions for $\mathbf{X}$ falling in that region only. In the previous decision tree models, their leaf models are always constant-valued (CART in \cite{breiman2017classification}), or linear regressions (SLRT in \cite{zheng2019partitioning}), or polynomial regressions (SUPPORT in \cite{chaudhuri1994piecewise}). Whereas we introduce the use of neural network which has the benefit that it can capture the nonlinearity and the interaction of the covariates.
                Their network structure can be less complicated than that of the splitting neural network $\hat{f}_{\text{split}}$ used in the internal nodes. It will be less difficult to model the data in leaf nodes than in internal nodes since the former may exhibit a smaller degree of heterogeneity. Besides, to avoid overfitting, we adopt necessary regularization strategies such as early stopping in all neural network training. % First, we divide data into the training set and validation set. We train our neural network model on the training set and calculate the root mean squared prediction errors of the validation set. When the fitting errors begin to increase, we terminate the optimization. 
                
                For different leaf regions, we adopt the same training scheme. Some works suggest to train $\hat{f}_m(\mathbf{X})$ and $\hat{\sigma}_m(\mathbf{X})$ simultaneously in one network with a gradient clipping strategy \cite{lakshminarayanan2017simple}, while some others train one of them with another fixed alternately and iteratively \cite{skafte2019reliable}. We adopt the latter one. %Different from some recent suggestions on how to train $\hat{f}_m(\mathbf{X})$ and $\hat{\sigma}_m(\mathbf{X})$, we take much simpler training steps. 
                First, $\hat{f}_m(\mathbf{X})$ is trained as usual with the mean square error loss. Then $\hat{f}_m(\mathbf{X})$ is fixed and $\hat{\sigma}_m(\mathbf{X})$ is trained with the negative log-likelihood of Gaussian whose mean and variance are $\hat{f}_m(\mathbf{X})$ and $\hat{\sigma}_m^2(\mathbf{X})$, and vice versa:
                \begin{align}
                        & \min_{\Theta^f_m,\Theta^\sigma_m}\sum_{\mathbf{X}\left(i\right)\in\mathcal{R}_m}-\log \mathbb{P}\left(\mathbf{y}\left(i\right) \mid \mathbf{X}\left(i\right)\right)= \nonumber \\
                        & \sum_{\mathbf{X}\left(i\right)\in\mathcal{R}_m}\frac{\log \hat{\sigma}_m^{2}\left(\mathbf{X}\left(i\right)\right)}{2}+\frac{\left(\mathbf{y}\left(i\right)-\hat{f}_m\left(\mathbf{X}\left(i\right)\right)\right)^{2}}{2 \hat{\sigma}_m^{2}\left(\mathbf{X}\left(i\right)\right)}+\text{constant},
                \end{align}
                where $\Theta^f_m,\Theta^\sigma_m$ are the learnable parameters of $\hat{f}_m(\mathbf{X})$ and $\hat{\sigma}_m(\mathbf{X})$, respectively.  %The reason why we adopt a simpler training scheme is again that the data in leaf regions may not exhibit heterogeneity and will be less difficult to model. Moreover, this will reduce the computational cost significantly.
                Finally, USNRT's prediction and uncertainty estimation are $\hat{f}(\mathbf{X}) = \sum_{m=1}^{M}\hat{f}_m(\mathbf{X})\mathbb{I}_{\left\{\mathbf{X} \in \mathcal{R}_{m}\right\}}$ and $\hat{\sigma}\left(\mathbf{X}\right) = \sum_{m=1}^{M}\hat{\sigma}_m\left(\mathbf{X}\right)\mathbb{I}_{\left\{\mathbf{X} \in \mathcal{R}_{m}\right\}}$.
                
                To further illustrate the necessity of the proposed method, we will compute the leaf region-specific residuals' variance using all $\left( \mathbf{y}(i)-\hat{f}_m(\mathbf{X}(i)) \right)^2$ when $\mathbf{X}(i)\in\mathcal{R}_m$. In Fig. \ref{Uncertainty_difference} in the experimental section, we show that the variances obtained are quite different or heterogeneous across leaf regions, implying the difficulty of quantifying uncertainty using one model for the whole feature space.

                \subsection{Computational Complexity}
                
        In tree construction, the two most time-consuming parts are searching for the best splits and training neural networks. The cost of training neural networks depends on the architecture. We use $O\left(end\times c(r)\right)$ to denote this cost. Here $e$ is the number of epochs, $n$ is the sample size in a node, $d$ is the feature dimension, $r$ represents the architecture, and $c$ is a function of $r$. Now the most crucial thing is that as the tree grows to be deeper, we need to train more neural networks. The number of neural networks increases exponentially, but meanwhile, the sample size in the node decreases exponentially. Consequently, the computational cost at every level of the tree remains the same. So, the total computational cost is proportional to the tree depth: $O\left(heNd\times c(r)\right)$, where $h$ is the tree depth and $N$ is the total sample size. In the experiments, 10 leaf nodes are enough for producing good performance, implying $h\approx3$. On some datasets, $h\approx2$ if the total sample size is not large.
        
        The searching for the best split takes $O(dn^2)$ computational cost because we perform Levene's test $O(dn)$ times. This can be reduced to $O(dn/s)$ times if we set a step size $s$ in searching for the split value $a$. The cost now is $O(dn^2/s)$ with a very small constant in this $O$. Furthermore, the cost at every level of the tree will decrease exponentially because of the $n^2$ term.
        In our experiments, this part costs much less than training neural networks does and can be ignored.
                
        %	\subsection{USNRT Ensemble}
        %	
        %	It is a familiar strategy to use ensemble learning method to improve the predictive ability of models. % One of the most commonly used ensemble methods is bagging, like random forests in \cite{breiman2001random}. The benifits of bagging are that it can boost the model performance as well as calculate in parallel computation. 
        %	In a similar way as in \cite{lakshminarayanan2017simple}, we use our USNRT as a base predictor to construct the deep ensemble of multiple USNRTs. Unlike in traditional ensemble, bagging is not used here. With different random initializations of neural networks, we can establish a set of USNRTs $T_1, T_2, \dots, T_J$ and obtain $J$ mean predictors $\hat{f}_1(\mathbf{X}), \hat{f}_2(\mathbf{X}), \dots, \hat{f}_J(\mathbf{X})$ and $J$ variance predictors $\hat{\sigma}^{2}_1\left(\mathbf{X}\right), \hat{\sigma}^{2}_2\left(\mathbf{X}\right), \dots, \hat{\sigma}^{2}_J\left(\mathbf{X}\right)$. The output of USNRT Ensemble is the mixture of $J$ Gaussian distributions. In experiments, $5$ USNRTs are ensembled and are enough to produce good performances.
        %%	\begin{equation}
        %%		\hat{f}_{\text{DN-USNRT}}(\mathbf{X}) = \frac{1}{J}\sum_{j = 1}^{J}\hat{f}_j(\mathbf{X}),
        %%	\end{equation}
        %%	and
        %%	\begin{equation}
        %%		\hat{\sigma}^{2}_{\text{DN-USNRT}}(\mathbf{X}) = \frac{1}{J}\sum_{j = 1}^{J}\hat{\sigma}^{2}_j(\mathbf{X}).
        %%	\end{equation}
        
        \section{Experiment of Aleatory Uncertainty}
        
        In this section, we evaluate USNRT on the task of aleatory uncertainty estimation and conduct the comparison to some competing models. We also analyze the stability of USNRT's performance with respect to the variation of hyper-parameters.
        
        \subsection{Data and Settings}
        
        \textbf{Datasets.} We collect 17 datasets of regression tasks from the UCI Machine Learning Repository \cite{Dua:2019}. The sample sizes range from 10,000 to 287,999, and the feature dimensions range from 4 to 276. In some datasets, the categorical variables are also included. These datasets are extensive and representative, and include some large-scale ones, making the experimental conclusions more convincing. For every dataset, we have a short name and a full name. The information of these datasets is detailed in Table \ref{dataset_info}. 
        
        \begin{table}[tbp]
                \centering
                \caption{Dataset information.}
                \label{dataset_info}
                \resizebox{\columnwidth}{!}{
                        \begin{tabular}{ccccc}
                                \toprule 
        \makecell{Dataset\\Name} & Full Name & \makecell{Sample\\Size} & \makecell{Feature\\Dimension} & \makecell{Categorical\\Variables} \\ \midrule
        Electrical & Electrical Grid Stability & 10,000 & 12 & 0\\
        Conditional & Conditional Based Maintenance & 11,933 & 15 & 0\\
        Appliances & Appliances Energy Prediction & 19,735 & 27 & 0\\
        Real-time & Real-time Election & 21,643 & 23 & 0\\
        Industry & Industry Energy Consumption & 35,040 & 9 & 3\\
        Facebook1 & Facebook Comment Volume 1 & 40,949 & 52 & 0\\
        Beijing & Beijing PM2.5 & 41,757 & 8 & 2\\
        Physicochemical & Physicochemical Properties & 45,730 & 9 & 0\\
        Traffic & Traffic Volume & 48,204 & 6 & 2\\
        Blog & Blog Feedback & 52,397 & 276 & 0\\
        Power & Power Consumption of T & 52,416 & 5 & 0\\
        Online & Online Video & 68,784 & 18 & 2\\
        Facebook2 & Facebook Comment Volume 2 & 81,312 & 52 & 0\\
        Year & Year Prediction MSD & 100,000 & 90 & 0\\
        Query & Query Analytics & 199,843 & 4 & 0\\
        GPU & GPU Kernel Performance & 241,600 & 14 & 4\\
        Wave & Wave Energy Converters & 287,999 & 48 & 0\\ \bottomrule
        
                \end{tabular}}
        \end{table}

                \textbf{Evaluation Settings.} For each dataset, we randomly select 80\% of the data for training and the rest for testing. With different randomness seeds, this splitting is done five times if the sample size is less than 100,000 and 1 time otherwise (the last four datasets in Table \ref{dataset_info}, for reducing computational burden). For every method, the evaluation metrics are averaged over these training/testing splits.
                 During the training of all neural network models, we choose 20\% from the training set as the validation set for early stopping, set the batch size to 64, set the maximum number of epochs to 1,000, and optimize using Adam \cite{kingma2014adam} with the learning rate 0.01. All continuous variables and the label are normalized to have sample mean 0 and sample variance 1. Categorical variables are converted to one-hot encodings. 
                
                \textbf{Specification of USNRT.} We denote the feature dimension before one-hot encoding as $d$. In USNRT, the splitting model $\hat{f}_{\text{split}}(\mathbf{X})$ in the internal nodes is set to have two hidden layers with sizes $[8d,4d]$. The prediction model $\hat{f}_{m}(\mathbf{X})$ and uncertainty estimation model $\hat{\sigma}_{m}^2(\mathbf{X})$ in the leaf nodes have two hidden layers too, with layer sizes $[4d,2d]$. For $\hat{f}_{\text{split}}(\mathbf{X})$ and $\hat{f}_{m}(\mathbf{X})$, the ReLU activation is used for hidden layers and the linear is used for the output layer. For $\hat{\sigma}_{m}^2(\mathbf{X})$, the Tanh activation is used for hidden layers and the Softplus is used for the output to ensure positiveness. For the two specific hyper-parameters of USNRT, we set the significance level $\alpha = 0.01$, and the minimum sample size in leaf nodes $N_{\min} = \max\{N_{\text{train}}/10, 1000\}$, where $N_{\text{train}}$ is the sample size of training data. This means that the constructed USNRT will have 10 leaf nodes at most, with about 3 levels (depth).% Moreover, we collect 5 USNRTs with different initializations to construct the USNRT Ensemble. 
                
                \textbf{Competing Models.}
                We compare USNRT to some models that estimate aleatory uncertainty only, including deep learning-based and traditional tree-based. They are Heteroskedastic Neural Network (HNN), Extra Trees, and Random Forest. They all produce one mean and one variance for each input for quantifying uncertainty, as USNRT does.
                In HNN, we build two neural networks to output the mean and the variance respectively. We train them alternately twice with negative log-likelihood loss, as suggested in \cite{skafte2019reliable}. Both the two networks have hidden layer sizes $[8d,4d]$, with ReLU and Tanh activations respectively. The output layer activations are linear and Softplus respectively. 
                For the Extremely Randomized Trees (ExTra) \cite{geurts2006extremely} and Random Forest \cite{breiman2001random}, we select their hyper-parameters with the negative log-likelihood loss on the validation set: the number of trees is in the range $\{50,100,150,200\}$; the tree depth is in $\{4,6,8,10,12\}$; and the proportion of features used for node splitting is in $\{0.3,0.5,0.7,0.9\}$. We use the implementations in the scikit-optimize package \cite{scikit-optimize}, in which the standard deviation for the prediction can be returned directly.
                %For the settings of Random Forest and ExTra, like the number of trees, the max depth, and the minimum numbles of samples on the leaves, we use cross-validation to select the optimal hyperparameters.

        \subsection{Evaluation Metrics}
        
        Given a testing dataset $\{(\mathbf{X}(i),\mathbf{y}(i))\}_{i\in\mathcal{I}_{te}}$, for any method that can output the mean and the variance for $\mathbf{y}(i)$ given $\mathbf{X}(i)$, we use three evaluation metrics to evaluate it. They are expected calibration error (ECE), tail-interval calibration error (TCE), and sharpness. Our definitions of calibration and sharpness follow the literature \cite{gneiting2007probabilistic,kuleshov2018accurate}.
        Denoting the mean and the variance given by the method as $\tilde{\mu}_i$ and $\tilde{\sigma}_i^2$, we first compute the predicted conditional $\tau_k$-quantile of $\mathbf{y}(i)$ given $\mathbf{X}(i)$ under the Gaussian assumption: 
        \begin{equation}\label{eqn:quantiles_Gaussian}
        \hat{q}_{i,k}=\tilde{\mu}_i+\tilde{\sigma}_i \Phi^{-1}(\tau_k),
        \end{equation}
        where $\Phi^{-1}(\cdot)$ is the inverse of the distribution function of standard Gaussian. We choose $\tau_k\in \Pi=\{0.01,0.02,\dots,0.99\}$, where $k=1,\dots,99$. With these predicted quantiles, we introduce the evaluation metrics as follows.
        
        \subsubsection{Expected Calibration Error (ECE)}
        In the regression problem, calibration means that the probability that the observed random variable is lower than the $\tau$-quantile equals the expected probability $\tau$. Here for $\tau_k$, the observed probability is 
        \begin{equation}
        \hat{P}^{\text{ece}}_k = \frac{1}{|\mathcal{I}_{te}|}\sum_{i\in\mathcal{I}_{te}}\mathbf{1}_{\{\mathbf{y}(i)<\hat{q}_{i,k}\}}.
        \end{equation}
        It is ideal if $\hat{P}^{\text{ece}}_k = \tau_k$. We compute the difference between them and take the average over all $k$ as ECE:
        \begin{equation}
                \text{ECE} = 100\times \frac{1}{K}\sum_{k=1}^{K}|\hat{P}^{\text{ece}}_k - \tau_k|. 
        \end{equation}
        The $K = 99$ probability levels in $\Pi$ spread the interval $(0,1)$, hence ECE evaluates the full predicted distribution.
        
        \subsubsection{Tail-Interval Calibration Error (TCE)}
        In many uncertainty quantification tasks, the prediction invertals are required to be generated. The calibration of interval means that the probability that the observed random variable is located in the interval equals the expected probability,  for example, 90\%. Usually the predicted 5\%- and 95\%-quantiles are used to form the interval in this case. Now for $\tau_k<0.5$, we construct the interval as $[\hat{q}_{i,k},\hat{q}_{i,100-k}]$ and the observed probability is 
        \begin{equation}
        \hat{P}^{\text{tce}}_k = \frac{1}{|\mathcal{I}_{te}|}\sum_{i\in\mathcal{I}_{te}}\mathbf{1}_{\{\hat{q}_{i,k}<\mathbf{y}(i)<\hat{q}_{i,100-k}\}}.
        \end{equation}
        The corresponding expected probability is $1-2\tau_k$. Ideally, they are equal. Our TCE computes their difference and takes the average over some $k$:
        \begin{equation}\label{eqn:TCE}
                \text{TCE} = 100\times \frac{1}{|\Pi'|}\sum_{\tau_k\in\Pi'}|\hat{P}^{\text{tce}}_k - (1-2\tau_k)|,
        \end{equation}
        where $\Pi'=\{0.05,0.1,0.15,0.2\}$. This makes TCE the average calibration error of 90\%, 80\%, 70\%, and 60\% intervals.
        
        %Therefore, we construct predicted intervals to see whether this calibration property would still hold. We use tow-side equal tail interval to define the $q$ nominal level interval as $\mathbb{I}_{q} = [\mathbf{y}_{\frac{1-q}{2}}, \mathbf{y}_{\frac{1+q}{2}}]$. The empirical coverage for $q$ nominal level interval is defined as 
        %\begin{equation}
        %	\hat{q\times100\%} = \frac{1}{N}\sum_{i=1}^{N}\mathbf{1}_{\mathbf{y}_i\in\hat{\mathbb{I}}_{q,i}}. 
        %\end{equation}
        %And we can calculate miscoverage at the $q$ nominal level as follows:
        %\begin{equation}
        %	q\times100\%\text{-interval ECE} = |\hat{q\times100\%} - q|.
        %\end{equation}
        %Here we set $q = \{0.1, 0.2, \cdots, 0.9\}$ and visualize the relationship between $q\times100\%\text{-interval ECE}$ and the nominal level $q$. If the method has well-calibrated interval coverage, its line is close to the zero horizontal line.
        
        %In some areas, people take interest in 90\% intervals for decision making. Therefore, we emphasize the 90\% interval calibration comparison. As defined above, the estimated 90\% interval is $\hat{\mathbb{I}}_{0.9} = [\hat{q}_{0.05}, \hat{q}_{0.95}]$.The empirical coverage for 90\% intervals is $\hat{90\%} = \frac{1}{N}\sum_{i=1}^{N}\mathbf{1}_{\mathbf{y}_i\in\hat{\mathbb{I}}_{0.9,i}}$. The miscoverage at the nominal 90\% level is $90\%\text{-interval ECE} = |\hat{90\%} - 0.9|$.
        %}
        
        \subsubsection{Sharpness}
        When calibration is one side of evaluating uncertainty estimation, sharpness is another side to complement the evaluation. Supposing a model outputs $\tilde{\mu}_i,\tilde{\sigma}_i^2$ such that $\mathcal{N}(\tilde{\mu}_i,\tilde{\sigma}_i^2)$ is the unconditional distribution of $\{\mathbf{y}(i)\}_{i\in\mathcal{I}_{te}}$, we may obtain nearly perfect calibration although it's not what we want. Sharpness measures the length of the intervals construced in TCE and makes narrow intervals preferred. One needs to find a good balance between calibration and sharpness. In our experiment, the quantiles are computed under the Gaussian assumption as in Equation \eqref{eqn:quantiles_Gaussian}. So, the length of the interval is proportional to the standard deviation $\tilde{\sigma}_i$. Instead, we can measure $\tilde{\sigma}_i$ directly in our sharpness metric:
        \begin{equation}
                \text{sharpness} =  100\times \frac{1}{|\mathcal{I}_{te}|}\sum_{i\in\mathcal{I}_{te}} \tilde{\sigma}_i.
        \end{equation}
        It is an alternative to the commonly-used interval sharpness $100\times \frac{1}{|\mathcal{I}_{te}|}\sum_{i\in\mathcal{I}_{te}} |\hat{q}_{i,100-k}-\hat{q}_{i,k}|$ for any $k$ because of Equation \eqref{eqn:quantiles_Gaussian}.
        
        %However, if we compare the interval sharpness under the nominal levels straightly, it seems unfair and hard to tell which method is better. Because a method may get a lower interval sharpness with the sacrifice of interval calibration and overconfidence. To demonstrate the trade-off between interval calibration and sharpness, we visualize the relationship between sharpness $q\times 100\%\text{-interval width}$ and empirical levels $\hat{q\times100\%}$. If a method has sharp calibrated intervals, its line is at a lower position.
        
        %Besides, 90\%-interval results are followed with interest, we also report the 90\%-interval sharpness which is defined as $90\%\text{-interval width} = \frac{1}{N}\sum_{i=1}^{N}|\hat{\mathbb{I}}_{0.9,i}|$.
        
        \subsection{Comparison of Performance}
        
        \textbf{ECE.} Table \ref{ece_result_table} shows ECE results of the four models on 17 datasets. As we can see, USNRT performs the best among all on 15 datasets out of 17. This is indeed a significantly superior performance. Moreover, on some datasets such as Conditional, Real-time, Online, and Wave, the improvement of USNRT's result over the others' best is close to or larger than 50\% (the last column). Overall, the second-place winner is HNN, which performs better than the two tree-based models. 
        
        \begin{table}[t]
                \centering
                \caption{\textbf{ECE} results. The bold number represents the best performance, and the second best is marked by *. USNRT performs the best on 15 datasets out of 17. The last column shows the percentage decrease of USNRT's result over the best of other three models.}
                \label{ece_result_table}
                \resizebox{\columnwidth}{!}{
                        \begin{tabular}{cccccc}
                                \toprule
				Dataset & HNN & \makecell{Extra\\Trees} & \makecell{Random\\Forest} & USNRT & \makecell{Percentage\\Decrease} \\ \midrule
				Electrical & \textbf{3.74} & 9.38 & $^*$6.09 & 6.92 & -84.9\% \\
				Conditional & $^*$5.91 & 15.80 & 11.49 & \textbf{2.14} & 63.7\% \\
				Appliances & $^*$8.68 & 14.68 & 13.87 & \textbf{5.59} & 35.6\% \\
				Real-time & $^*$13.79 & 23.11 & 22.90 & \textbf{2.34} & 83.1\% \\
				Industry & $^*$3.44 & 8.87 & 11.53 & \textbf{2.49} & 27.7\% \\
				Facebook1 & $^*$9.81 & 14.33 & 12.23 & \textbf{7.80} & 20.5\% \\
				Beijing & $^*$3.73 & 7.28 & 6.04 & \textbf{3.33} & 10.7\% \\
				Physicochemical & $^*$3.07 & 5.51 & 6.03 & \textbf{2.72} & 11.3\% \\
				Traffic & 3.74 & \textbf{3.45} & $^*$3.51 & 3.67 & -6.6\% \\
				Blog & $^*$15.43 & 17.79 & 17.02 & \textbf{12.40} & 19.7\% \\
				Power & 2.26 & 2.53 & $^*$2.09 & \textbf{2.04} & 2.3\% \\
				Online & $^*$1.91 & 6.41 & 5.87 & \textbf{0.89} & 53.5\% \\
				Facebook2 & $^*$11.44 & 14.67 & 12.61 & \textbf{9.13} & 20.2\% \\
				Year & $^*$3.49 & 7.63 & 7.05 & \textbf{3.35} & 3.9\% \\
				Query & $^*$1.62 & 12.24 & 9.56 & \textbf{1.13} & 30.1\% \\
				GPU & $^*$3.04 & 8.17 & 5.95 & \textbf{2.32} & 23.5\% \\
				Wave & 14.72 & 9.45 & $^*$5.98 & \textbf{1.52} & 74.6\% \\ \bottomrule
                \end{tabular}}
        \end{table}
        
        %%%%%%%%%%%%%%%%%%%%%%%%%%%%%%%%%%%%%%%%%%%%%%%%%%%%%%%%%%%%%%
        
        \textbf{TCE.} Table \ref{ece90_result_table} shows TCE results of the four models on 17 datasets. As we can see, USNRT performs the best among all on 11 datasets out of 17, and  performs the second best or above on 15 datasets out of 17. This performance is excellent, considering the difficulty of tail-side calibration. Moreover, on datasets Appliances, Real-time, Beijing, Online, Year, and Wave, the improvement of USNRT's result over the others' best is close to or larger than 50\% (the last column). Overall, the second-place winner is again HNN, which wins the best on 5 datasets out of 17.
        
        \begin{table}[t]
                \centering
                \caption{\textbf{TCE} results. The bold number represents the best performance, and the second best is marked by *. USNRT performs the best on 11 datasets out of 17, while HNN does on 5. The last column shows the percentage decrease of USNRT's result over the best of other three models.}
                \label{ece90_result_table}
                \resizebox{\columnwidth}{!}{
                        \begin{tabular}{cccccc}
                                \toprule 
				Dataset & HNN & \makecell{Extra\\Trees} & \makecell{Random\\Forest} & USNRT & \makecell{Percentage\\Decrease} \\ \midrule
				Electrical & \textbf{7.02} & 19.33 & $^*$14.18 & 19.92 & -183.8\% \\
				Conditional & $^*$2.39 & 24.47 & 20.50 & \textbf{1.57} & 34.3\% \\
				Appliances & $^*$11.24 & 19.25 & 18.76 & \textbf{6.33} & 43.7\% \\
				Real-time & $^*$14.96 & 24.11 & 23.71 & \textbf{2.55} & 82.9\% \\
				Industry & \textbf{3.36} & 20.14 & 20.71 & $^*$3.64 & -8.5\% \\
				Facebook1 & $^*$12.53 & 18.78 & 15.29 & \textbf{9.29} & 25.9\% \\
				Beijing & $^*$4.29 & 11.77 & 8.86 & \textbf{2.23} & 48.0\% \\
				Physicochemical & \textbf{2.47} & 10.09 & 10.87 & $^*$2.56 & -3.6\% \\
				Traffic & 7.59 & \textbf{6.62} & $^*$6.89 & 7.29 & -10.1\% \\
				Blog & $^*$19.12 & 21.46 & 20.63 & \textbf{14.21} & 25.7\% \\
				Power & $^*$2.52 & 7.06 & 6.66 & \textbf{2.29} & 9.3\% \\
				Online & $^*$1.56 & 14.54 & 12.84 & \textbf{0.92} & 41.4\% \\
				Facebook2 & $^*$16.41 & 20.04 & 17.24 & \textbf{11.90} & 27.5\% \\
				Year & $^*$3.54 & 10.95 & 9.90 & \textbf{1.63} & 53.9\% \\
				Query & \textbf{0.88} & 22.36 & 19.21 & $^*$1.31 & -50.0\% \\
				GPU & \textbf{2.61} & 17.65 & 13.75 & $^*$2.98 & -14.2\% \\
				Wave & 22.38 & 18.98 & $^*$14.24 & \textbf{2.73} & 80.9\% \\ \bottomrule
                \end{tabular}}
        \end{table}
        
        \textbf{Sharpness.} Table \ref{sharpness_result_table} shows sharpness results of the four models on 17 datasets. We can find that USNRT performs the best on 8 datasets out of 17, and performs the second best or above on 15 datasets out of 17, indicating that USNRT is surprisingly doing well on both calibration and sharpness. Besides, HNN and Random Forest give acceptable performance on sharpness.
        
        \begin{table}[t]
                \centering
                \caption{\textbf{Sharpness} results. The bold number represents the best performance, and the second best is marked by *. USNRT performs the best on 8 datasets out of 17, while HNN does on 4 and Random Forest does on 5.}
                \label{sharpness_result_table}
                %\resizebox{\columnwidth}{!}{
                	\footnotesize
                        \begin{tabular}{ccccc}
                                \toprule 
				Dataset & HNN & \makecell{Extra\\Trees} & \makecell{Random\\Forest} & USNRT \\ \midrule
				Electrical & \textbf{11.16} & 58.81 & 48.42 & $^*$13.71 \\
				Conditional & \textbf{3.56} & 37.06 & 24.36 & $^*$8.56 \\
				Appliances & $^*$64.03 & 75.59 & 72.23 & \textbf{54.51} \\
				Real-time & 2.01 & $^*$1.81 & \textbf{1.74} & 2.84 \\
				Industry & \textbf{1.26} & 6.82 & 4.09 & $^*$1.83 \\
				Facebook1 & 27.96 & 21.96 & $^*$18.03 & \textbf{17.46} \\
				Beijing & $^*$64.78 & 74.93 & 67.41 & \textbf{62.19} \\
				Physicochemical & $^*$60.58 & 79.79 & 67.73 & \textbf{60.45} \\
				Traffic & 96.61 & 95.94 & \textbf{95.04} & $^*$95.50 \\
				Blog & 56.60 & $^*$27.46 & \textbf{24.81} & 35.62 \\
				Power & 70.93 & 79.00 & \textbf{69.37} & $^*$70.32 \\
				Online & $^*$6.12 & 12.54 & 9.38 & \textbf{5.08} \\
				Facebook2 & 34.61 & 21.45 & \textbf{17.83} & $^*$17.97 \\
				Year & $^*$74.22 & 90.95 & 86.08 & \textbf{71.81} \\
				Query & $^*$7.79 & 21.76 & 10.97 & \textbf{7.33} \\
				GPU & 3.69 & 4.97 & $^*$3.40 & \textbf{2.75} \\
				Wave & \textbf{0.01} & 4.73 & 4.07 & $^*$0.05 \\ \bottomrule
                \end{tabular}%}
        \end{table}
        
        Overall, the performance improvements of USNRT over the three existing models are significant. Given the extensiveness of the datasets we adopt (including large-scale ones), we can conclude that USNRT is indeed a better method for the type of data considered in this paper. Intuitively, there must be some intrinsic reason for the performance improvements. Could it be attributed to the discovery of uncertainty heterogeneity? We will explore this later in the visualization and interpretation section.
        %{\color{red} It is notable that among all competing methods, Deep Ensemble is the most computationally costly because 5 HNNs need to be trained. Even for training one HNN, the computational cost is higher than or close to that of training an USNRT. Because the cost of training an HNN is equivalent to that of training four ordinary neural networks with the same size, while in USNRT, it is about three or at most four in our setting.}
        
        \subsection{Stability Analysis}
        
        \begin{figure*}[t]
                \centering
                \begin{subfigure}[t]{0.315\linewidth}
                        \centering
                        \includegraphics[width=\linewidth]{{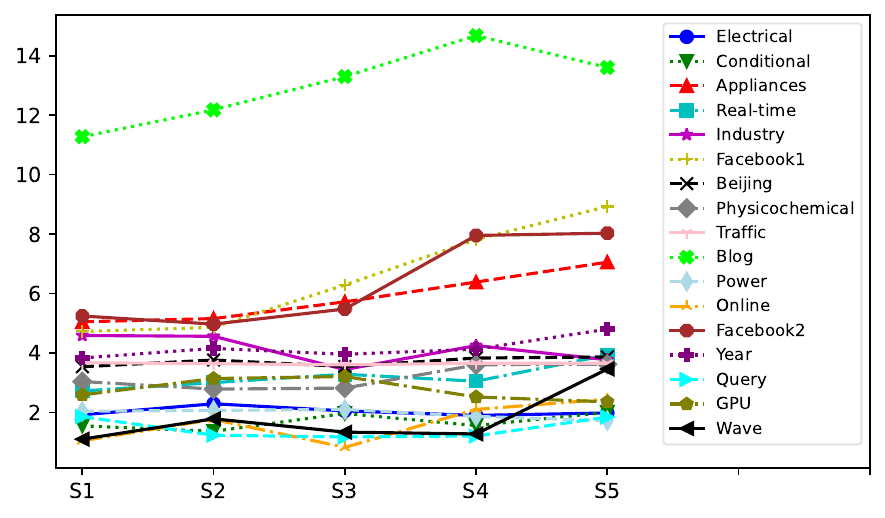}}
                        \caption{{\small ECE versus Network Structure}}
                \end{subfigure}\hspace{0.7em}
                \centering
                \begin{subfigure}[t]{0.315\linewidth}
                        \centering
                        \includegraphics[width=\linewidth]{{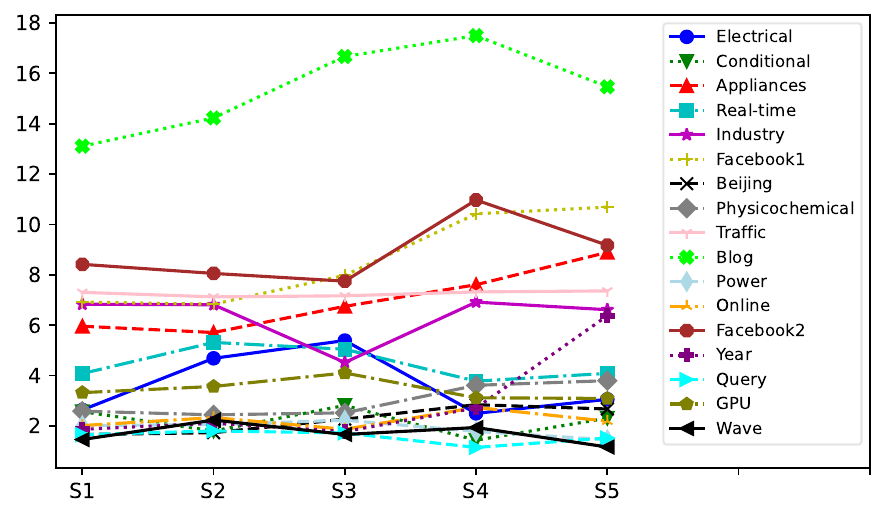}}
                        \caption{{\small TCE versus Network Structure}}
                \end{subfigure}\hspace{0.7em}
                \centering
                \begin{subfigure}[t]{0.315\linewidth}
                        \centering
                        \includegraphics[width=\linewidth]{{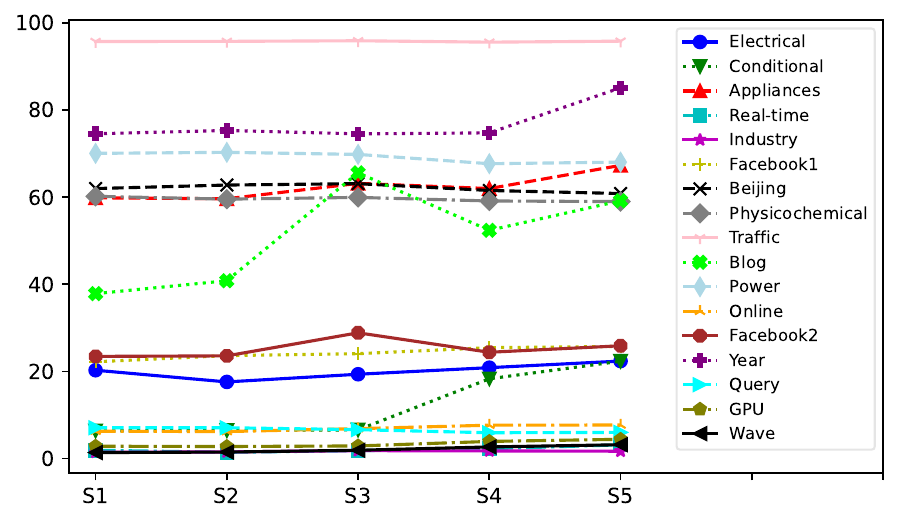}}
                        \caption{{\small Sharpness versus Network Structure}}
                \end{subfigure}\\
                \centering
                \begin{subfigure}[t]{0.315\linewidth}
                        \centering
                        \includegraphics[width=\linewidth]{{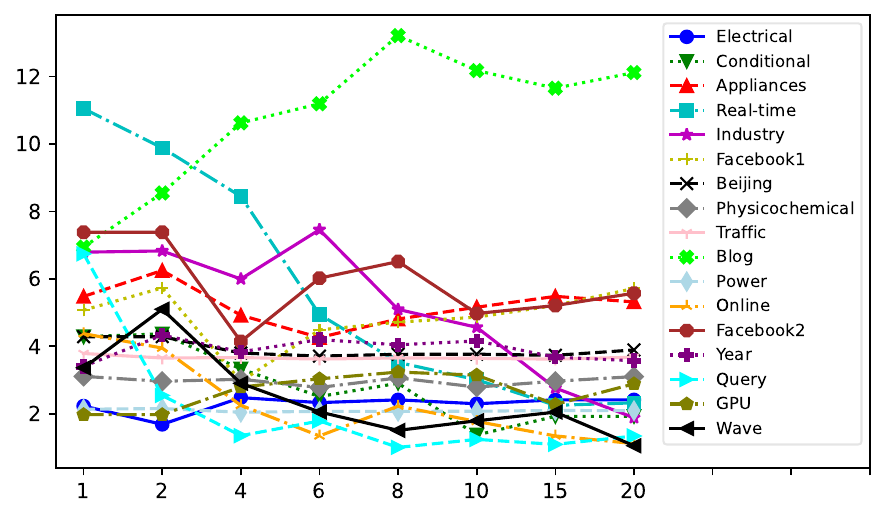}}
                        \caption{{\small ECE versus No. of Leaves $N_{\text{leaves}}$}}
                \end{subfigure}\hspace{0.7em}
                \centering
                \begin{subfigure}[t]{0.315\linewidth}
                        \centering
                        \includegraphics[width=\linewidth]{{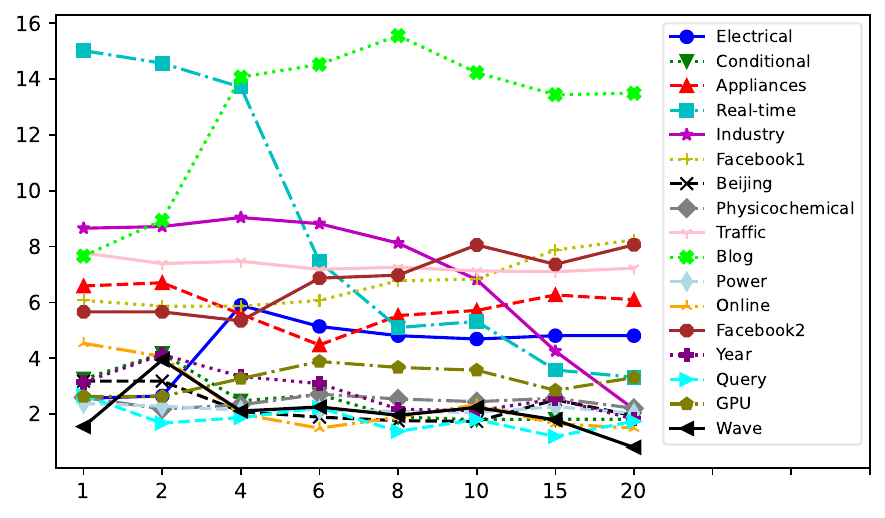}}
                        \caption{{\small TCE versus No. of Leaves $N_{\text{leaves}}$}}
                \end{subfigure}\hspace{0.7em}
                \centering
                \begin{subfigure}[t]{0.315\linewidth}
                        \centering
                        \includegraphics[width=\linewidth]{{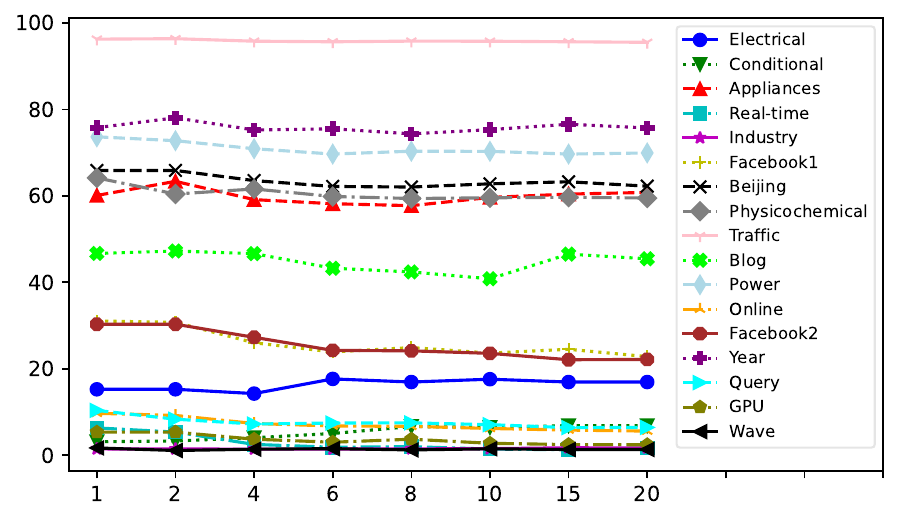}}
                        \caption{{\small Sharpness versus No. of Leaves $N_{\text{leaves}}$}}
                \end{subfigure}\\
                \centering
                \begin{subfigure}[t]{0.315\linewidth}
                        \centering
                        \includegraphics[width=\linewidth]{{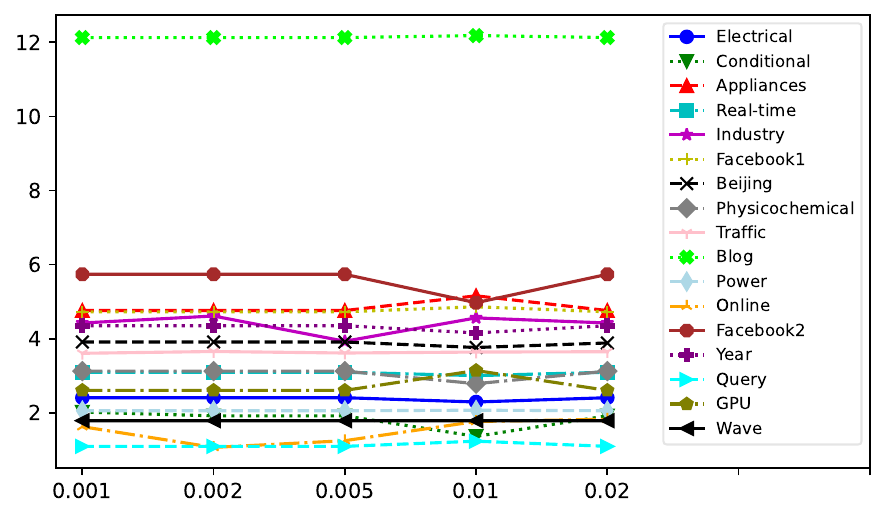}}
                        \caption{{\small ECE versus Significance Level $\alpha$}}
                \end{subfigure}\hspace{0.7em}
                \centering
                \begin{subfigure}[t]{0.315\linewidth}
                        \centering
                        \includegraphics[width=\linewidth]{{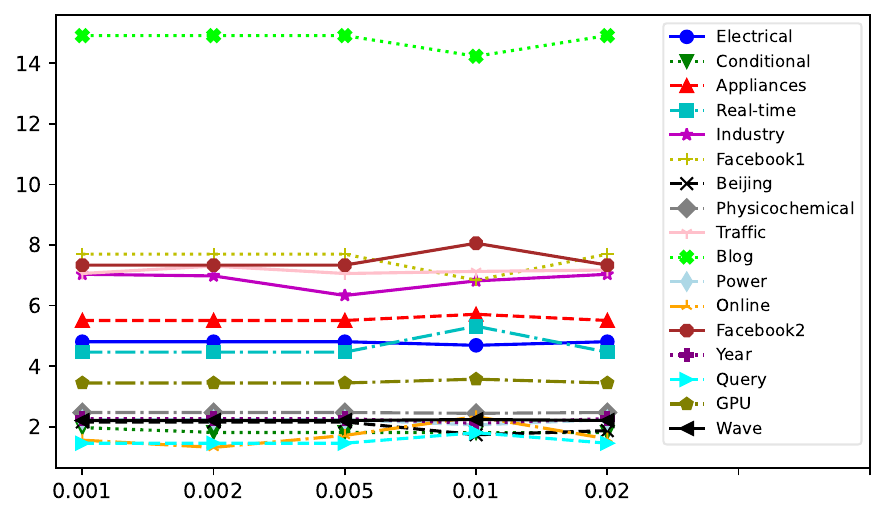}}
                        \caption{{\small TCE versus Significance Level $\alpha$}}
                \end{subfigure}\hspace{0.7em}
                \centering
                \begin{subfigure}[t]{0.315\linewidth}
                        \centering
                        \includegraphics[width=\linewidth]{{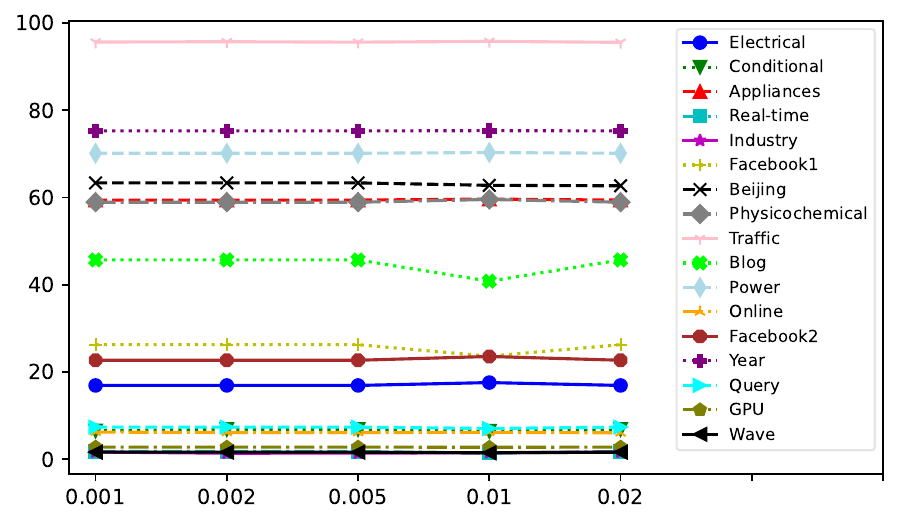}}
                        \caption{{\small Sharpness versus Significance Level $\alpha$}}
                \end{subfigure}
                \caption{The performance versus varying hyper-parameters on every dataset. (a)--(c): five different network structure combinations are in the $x$-axis, and from S1 to S5, the networks become larger and larger. One can find small networks are good enough for producing good ECE and TCE results. (d)--(f): the maximum number of leaf nodes $N_{\text{leaves}}$ varies in $\{1,2,4,6,8,10,15,20\}$ in the $x$-axis. On some datasets such as Conditional, Real-time, Industry, Online, Query, and Wave, ECE and TCE decrease substantially as $N_{\text{leaves}}$ increases. On other datasets except Blog, ECE and TCE decrease moderately or remain stable. (g)-(i): the significance level $\alpha$ varies in $\{1,2,5,10,20\}\times 10^{-3}$ and the performance remains stable.}
                \label{usnrt_stability}
        \end{figure*}
        
        In this subsection, we show that USNRT's performance is stable with respect to the changing hyper-parameters. The structure of neural networks in internal nodes and the structure in leaf nodes are of course important hyper-parameters. Besides, USNRT has two additional hyper-parameters used in seeking the best split: the significance level $\alpha$ and the minimum sample size in leaf nodes $N_{\text{min}}$.% Generally, $\alpha$ can be set to 0.01 without change, as in usual statistical tests. So, we will not analyze it in the following.
        
        For neural network structures, we change the hidden layers and consider five different combinations of the two structures in internal nodes and leaf nodes: $[4d, 2d]+[4d, 2d]$, $[8d, 4d]+[4d, 2d]$, $[16d, 8d]+[8d, 4d]$, $[8d, 16d, 4d]+[4d, 8d, 2d]$, and $[16d, 16d, 8d]+[8d, 8d, 4d]$. They make the networks more and more complex. We implement USNRT with them and plot the corresponding varying performance using the three metircs, as shown in Fig. \ref{usnrt_stability} (a)--(c), where each line corresponds to a dataset. One can see that larger networks do not necessarily lead to better ECE or TCE. Indeed, $[4d, 2d]+[4d, 2d]$ or $[8d, 4d]+[4d, 2d]$ (S1, S2 in the horizontal axis) is enough for good performance, and they have lower computational costs. The sharpness results are quite stable with respect to the changing structures. Conclusively, there is no need to carefully pick neural network structures.
        
        %Because our USNRT models include neural networks and a tree structure, there is a need to consider how to determine the number of hidden layers, the number of neurons in each layer, and the minimum sample size $N_{\min}$ in the leaf nodes (important hyper-parameter to determine the tree structure). USNRT needs two neural networks, the splitting model and the leaf model. Here, we consider five combinations to test the stability of USNRT against varying network structure. We use $[8, 4]+[4, 2]$ to denote that the spliting model has hidden layer sizes $\{8N_{\text{dim}}, 4N_{\text{dim}}\}$ and the leaf model has hidden layer sizes $\{4N_{\text{dim}}, 2N_{\text{dim}}\}$, where $N_{\text{dim}}$ is the input dimension. All the five combinations are $[4, 2]+[4, 2]$, $[8, 4]+[4, 2]$, $[16, 8]+[8, 4]$, $[8, 16, 4]+[4, 8, 2]$, and $[16, 16, 8]+[8, 8, 4]$. For the minimun sample size in the leaf nodes, we define $N_{\min} = \max\{N_{\text{train}}/N_{\text{split}}, 1000\}$, where $N_{\text{split}}$ varies in $\{1, 3, 5, 7, 10, 15, 20\}$. $N_{\text{split}}$ represents the maximum possible number of leaf nodes. When analyzing network structures, we fix $N_{\text{split}}=10$. When evaluating the influence of  $N_{\text{split}}$, we fix the network structure to be $[8, 4]+[4, 2]$. 
        
        Similarly, we study the effect of the minimum sample size in leaf nodes $N_{\text{min}}$ on the performance. For better illustration, we define $N_{\min} = \max\{N_{\text{train}}/N_{\text{leaves}}, 1000\}$, where $N_{\text{train}}$ is the size of training dataset, and $N_{\text{leaves}}$ will be the number of leaf nodes at most. We change $N_{\text{leaves}}$ in $\{1,2,4,6,8,10,15,20\}$ and plot the corresponding varying performance in Fig. \ref{usnrt_stability} (d)--(f). We can find that on some datasets such as Conditional, Real-time, Industry, Online, Query, and Wave, ECE and TCE results decrease substantially as $N_{\text{leaves}}$ increases. On other datasets except Blog, ECE and TCE decrease moderately or remain stable. This may be because different datasets have different degrees of heterogeneity. The sharpness results are quite stable again. So, setting $N_{\text{leaves}}$ to 8 or 10 will be an appropriate choice in moderation, and has acceptable computational cost as well.
        
        %where RMSE and ECE versus network structure and $N_{\text{split}}$ for 12 datasets are displayed. We can see that in most datasets, the network structure and $N_{\text{split}}$ have little effect on the prediction accuracy or RMSE, showing excellent stability. The change of network structure has a small fluctuation effect on uncertainty. The overall ECEs are stable while deeper model (the fifth structure) may produce worse results. With $N_{\text{split}}$ increasing, $N_{\min}$ is decreasing and there will be more leaf nodes in USNRT. Now ECE results become better and will remain stable after $N_{\text{split}}\ge 10$. Too large $N_{\text{split}}$ will significantly increase the computation cost and is not recommended.
        
        At last, we let the significance level $\alpha$ used in the splitting vary in $\{1,2,5,10,20\}\times 10^{-3}$, and plot the corresponding varying performance in Fig. \ref{usnrt_stability} (g)--(i). One can see that all evaluation metrics are stable when $\alpha$ varies, on every dataset. So, we can set $\alpha$ to 0.01 with no further change.
        
        \section{Experiment of Total Uncertainty}
        
        In this section, we evaluate USNRT Ensemble which consists of five USNRTs with different network initialization seeds, on the task of total uncertainty estimation (aleatory and epistemic). Different from aleatory uncertainty estimation, we conduct the comparison to different models (that can estimate total uncertainty) and use different evaluation metrics, which are more aligned with previous research works. 
        
        \subsection{Data, Settings, and Evaluation Metrics}
        
        \textbf{Datasets and Evaluation Settings.} The datasets and evaluation settings are the same as in the aleatory experiment. It is worth noting that the datasets used here are more extensive and representative than in some previous research works. We even include some large-scale datasets, as shown in Table \ref{dataset_info}. %Hence, our experimental conclusions will be more convincing.
        
        \textbf{Competing Models.} The specification of each USNRT in USNRT Ensemble is the same as in the aleatory experiment.
        We compare USNRT Ensemble to some recent popular models, including MC Dropout, Concrete Dropout, Deep Ensemble, and Deep Evidential Regression. They are either Bayesian or can output a mixture (or distribution) of Gaussian distributions. MC Dropout \cite{gal2016dropout} and Concrete Dropout \cite{gal2017concrete} are popular Bayesian approximation models recently. In our experiment, MC Dropout has hidden layer sizes $[8d,16d,4d]$, dropout rate $0.5$, ReLU/linear as hidden/output layer activation, and Mean Square Error as loss function. Concrete Dropout has hidden layer sizes $[8d,4d]$ and the dropout rate can be learned in the model (the official code released is used). After training, both the two models generate 1,000 predictions under the dropout mode for computing the mean and the variance. Deep Ensemble \cite{lakshminarayanan2017simple} trains 5 HNNs with different random initialization seeds, each of which has the same hyper-parameters as in the aleatory experiment. The five means and variances are aggregated to obtain a mixture of Gaussian distributions, which is also the case in USNRT Ensemble. For the Evidential method \cite{amini2020deep}, we again set hidden layer sizes to $[8d,4d]$ and their activation to ReLU (linear or Softplus for the output, suggested in \cite{amini2020deep}). For the most important  regularization coefficient, we first set $\lambda=0.01$ as suggested. We then change $\lambda$ to see if the outperformance of USNRT Ensemble changes.
        
        \textbf{Evaluation Metrics.} Following the prior line of works \cite{gal2016dropout,lakshminarayanan2017simple,amini2020deep}, we evaluate the total uncertainty estimation given by the five models described above with the evaluation metrics RMSE and Negative Log-Likelihood (NLL). All results are multiplied by 100 and then reported. %In addition, we aslo report the RMSE results given by Extra Trees and Random Forest, and compare them to USNRT Ensemble. We do this comparison because tree models are thought to perform better generally than neural networks on tabular data on prediction accuracy \cite{borisov2022deep}.
        
        \subsection{Comparison of Performance}
        
                \begin{table}[t]
                \centering
                \caption{\textbf{RMSE} results. The bold number represents the best performance, and the second best is marked by *. USNRT Ensemble performs the best on 10 datasets out of 17, and performs the second best or above on 14 out of 17.}
                \label{rmse_result_table}
                \resizebox{\columnwidth}{!}{
                        \begin{tabular}{cccccc}
                                \toprule
Dataset & \makecell{MC\\Dropout} & \makecell{Concrete\\Dropout} & \makecell{Deep\\Ensemble} & Evidential & \makecell{USNRT\\Ensemble} \\ \midrule
Electrical & 29.28 & 16.77 & \textbf{14.71} & $^*$16.51 & 18.55 \\
Conditional & 99.71 & 21.34 & \textbf{3.53} & $^*$4.36 & 9.61 \\
Appliances & 94.37 & 90.48 & $^*$80.33 & 86.62 & \textbf{79.19} \\
Real-time & 24.03 & 61.58 & $^*$6.75 & 12.87 & \textbf{6.05} \\
Industry & 10.73 & 3.68 & $^*$2.52 & \textbf{2.42} & 3.24 \\
Facebook1 & 82.54 & 390.74 & $^*$75.76 & 77.50 & \textbf{69.00} \\
Beijing & 79.92 & 76.42 & $^*$73.06 & 76.84 & \textbf{72.50} \\
Physicochemical & 79.48 & 78.75 & \textbf{62.98} & 71.53 & $^*$63.72 \\
Traffic & 97.80 & 99.82 & $^*$96.76 & 96.78 & \textbf{96.26} \\
Blog & $^*$94.82 & 313.93 & 95.53 & 102.72 & \textbf{76.58} \\
Power & 85.74 & 82.99 & $^*$80.19 & 82.26 & \textbf{79.67} \\
Online & 42.88 & 42.24 & $^*$10.73 & 13.01 & \textbf{10.00} \\
Facebook2 & 84.49 & 228.33 & $^*$72.55 & 76.78 & \textbf{71.57} \\
Year & 99.19 & 1049.22 & \textbf{80.20} & 83.38 & $^*$81.36 \\
Query & 35.49 & 32.89 & \textbf{16.41} & 21.74 & $^*$16.63 \\
GPU & 24.25 & 24.00 & $^*$5.89 & 6.94 & \textbf{3.89} \\
Wave & 7.52 & 3.38 & \textbf{0.01} & 0.31 & $^*$0.05 \\ \bottomrule
                \end{tabular}}
        \end{table}
        
        %%%%%%%%%%%%%%%%%%%%%%%%%%%%%%%%%%%%%%%%%%%%%%%%%%%%%%%%%%%%%%
        
                \begin{table}[t]
                \centering
                \caption{\textbf{NLL} results. The bold number represents the best performance, and the second best is marked by *. Over the 17 datasets, USNRT Ensemble performs the best on 6 datasets (6 for Evidential and 5 for Deep Ensemble), and performs the second best or above on 14 (10 for Evidential and 9 for Deep Ensemble).}
                \label{nll_result_table}
                \resizebox{\columnwidth}{!}{
                        \begin{tabular}{cccccc}
                                \toprule
Dataset & \makecell{MC\\Dropout} & \makecell{Concrete\\Dropout} & \makecell{Deep\\Ensemble} & Evidential & \makecell{USNRT\\Ensemble} \\ \midrule
Electrical & 27.11 & $^*$-89.39 & \textbf{-96.92} & -88.66 & -49.69 \\
Conditional & 29795.97 & 5.01 & \textbf{-239.25} & -162.63 & $^*$-183.49 \\
Appliances & 1181.69 & 103.00 & 64.49 & \textbf{10.59} & $^*$41.06 \\
Real-time & -40.40 & -95.67 & \textbf{-598.80} & $^*$-540.79 & -525.96 \\
Industry & -35.18 & -266.81 & \textbf{-415.01} & $^*$-407.99 & -378.31 \\
Facebook1 & 278.47 & inf & -150.55 & \textbf{-278.54} & $^*$-179.14 \\
Beijing & 318.08 & 85.81 & 73.48 & $^*$70.68 & \textbf{69.60} \\
Physicochemical & 610.50 & 111.01 & 81.60 & \textbf{64.18} & $^*$76.76 \\
Traffic & 755546.08 & 140.60 & $^*$137.47 & 140.80 & \textbf{136.80} \\
Blog & 542.35 & inf & -69.01 & \textbf{-195.01} & $^*$-126.42 \\
Power & 7480.57 & 100.35 & $^*$91.85 & 96.82 & \textbf{89.82} \\
Online & 15.48 & -111.76 & -214.04 & $^*$-215.23 & \textbf{-234.28} \\
Facebook2 & 407.42 & 212.48 & -145.25 & \textbf{-283.05} & $^*$-180.91 \\
Year & 7060.35 & inf & 99.19 & \textbf{94.87} & $^*$98.81 \\
Query & 78.01 & -166.92 & $^*$-202.46 & -178.14 & \textbf{-218.70} \\
GPU & 89.27 & -165.76 & $^*$-250.53 & -237.43 & \textbf{-283.45} \\
Wave & -44.61 & -183.99 & \textbf{-807.34} & -489.26 & $^*$-698.28 \\ \bottomrule
                \end{tabular}}
        \end{table}
        
        %%%%%%%%%%%%%%%%%%%%%%%%%%%%%%%%%%%%%%%%%%%%%%%%%%%%%%%%%%%%%%
        
                \begin{table}[t]
                \centering
                \caption{\textbf{Evidential RMSE} results with varying $\lambda$. The bold number represents the best performance, and the second best is marked by *. USNRT Ensemble performs the best on 14 datasets out of 17.}
                \label{evidential_rmse_result_table}
                \resizebox{\columnwidth}{!}{
                        \begin{tabular}{cccccc}
                                \toprule
Dataset & \makecell{Evidential\\$\lambda=0$} & \makecell{Evidential\\$\lambda=0.005$} & \makecell{Evidential\\$\lambda=0.01$} & \makecell{Evidential\\$\lambda=0.05$} & \makecell{USNRT\\Ensemble} \\ \midrule
Electrical & \textbf{16.34} & 16.62 & $^*$16.51 & 16.94 & 18.55 \\
Conditional & 4.70 & $^*$4.57 & \textbf{4.36} & 5.18 & 9.61 \\
Appliances & 86.14 & $^*$85.47 & 86.62 & 86.51 & \textbf{79.19} \\
Real-time & 30.37 & 22.27 & $^*$12.87 & 44.39 & \textbf{6.05} \\
Industry & 2.47 & \textbf{2.41} & $^*$2.42 & 2.56 & 3.24 \\
Facebook1 & $^*$74.44 & 77.70 & 77.50 & 78.54 & \textbf{69.00} \\
Beijing & $^*$75.62 & 76.36 & 76.84 & 78.32 & \textbf{72.50} \\
Physicochemical & $^*$71.41 & 71.85 & 71.53 & 71.85 & \textbf{63.72} \\
Traffic & $^*$96.73 & 96.82 & 96.78 & 96.91 & \textbf{96.26} \\
Blog & 1116631.33 & $^*$93.57 & 102.72 & 99.86 & \textbf{76.58} \\
Power & 82.09 & $^*$81.95 & 82.26 & 82.85 & \textbf{79.67} \\
Online & 13.35 & $^*$12.85 & 13.01 & 14.46 & \textbf{10.00} \\
Facebook2 & 73.76 & 74.20 & 76.78 & $^*$71.87 & \textbf{71.57} \\
Year & 83.84 & $^*$82.80 & 83.38 & 84.89 & \textbf{81.36} \\
Query & 29.49 & 27.51 & 21.74 & $^*$21.65 & \textbf{16.63} \\
GPU & 6.59 & $^*$6.12 & 6.94 & 6.31 & \textbf{3.89} \\
Wave & $^*$0.23 & 0.32 & 0.31 & 0.34 & \textbf{0.05} \\ \bottomrule
                \end{tabular}}
        \end{table}
        
                \begin{table}[t]
                \centering
                \caption{\textbf{Evidential NLL} results with varying $\lambda$. The bold number represents the best performance, and the second best is marked by *. USNRT Ensemble performs the best on 8 datasets out of 17.}
                \label{evidential_nll_result_table}
                \resizebox{\columnwidth}{!}{
                        \begin{tabular}{cccccc}
                                \toprule
Dataset & \makecell{Evidential\\$\lambda=0$} & \makecell{Evidential\\$\lambda=0.005$} & \makecell{Evidential\\$\lambda=0.01$} & \makecell{Evidential\\$\lambda=0.05$} & \makecell{USNRT\\Ensemble} \\ \midrule
Electrical & -85.47 & $^*$-87.01 & \textbf{-88.66} & -86.00 & -49.69 \\
Conditional & -152.35 & -157.47 & $^*$-162.63 & -145.06 & \textbf{-183.49} \\
Appliances & $^*$10.37 & \textbf{9.74} & 10.59 & 12.04 & 41.06 \\
Real-time & -542.03 & \textbf{-556.76} & -540.79 & $^*$-543.69 & -525.96 \\
Industry & -403.07 & -400.53 & \textbf{-407.99} & $^*$-407.03 & -378.31 \\
Facebook1 & \textbf{-282.58} & $^*$-280.99 & -278.54 & -276.24 & -179.14 \\
Beijing & $^*$69.91 & 70.57 & 70.68 & 72.89 & \textbf{69.60} \\
Physicochemical & \textbf{63.80} & 64.81 & $^*$64.18 & 69.15 & 76.76 \\
Traffic & $^*$137.64 & 139.90 & 140.80 & 144.93 & \textbf{136.80} \\
Blog & -155.48 & $^*$-199.20 & -195.01 & \textbf{-204.11} & -126.42 \\
Power & $^*$95.72 & 96.08 & 96.82 & 100.05 & \textbf{89.82} \\
Online & $^*$-216.43 & -214.45 & -215.23 & -211.76 & \textbf{-234.28} \\
Facebook2 & \textbf{-285.46} & -284.08 & -283.05 & $^*$-285.05 & -180.91 \\
Year & $^*$95.32 & 95.64 & \textbf{94.87} & 98.37 & 98.81 \\
Query & $^*$-179.23 & -177.09 & -178.14 & -170.57 & \textbf{-218.70} \\
GPU & $^*$-238.44 & -236.76 & -237.43 & -236.45 & \textbf{-283.45} \\
Wave & -489.05 & -485.53 & $^*$-489.26 & -488.41 & \textbf{-698.28} \\ \bottomrule
                \end{tabular}}
        \end{table}
        
         Table \ref{rmse_result_table} shows RMSE results of the five models on 17 datasets. As it shows, USNRT Ensemble performs the best on 10 datasets out of 17, and performs the second best or above on 14 out of 17, indicating the superior performance. Table \ref{nll_result_table} shows NLL results of the five models on 17 datasets. Although the performance of USNRT Ensemble is not as good as in the RMSE table, it still shows comparable performance over Deep Ensemble and Evidential. Next, we change the regularization coefficient $\lambda$ in Evidential from $0.01$ to other values and see its performance variation and the comparison  with USNRT Ensemble. Table \ref{evidential_rmse_result_table} and \ref{evidential_nll_result_table} show the RMSE and NLL results respectively. One can find that USNRT Ensemble outperforms Evidential with various $\lambda$ substantially.

        \section{Visualization and Interpretation}
        
        Through visualizing the learning results of the proposed USNRT, we try to understand what we have actually learned from the data and whether they are coherent with the motivation of this paper: the uncertainty heterogeneity, or not. Indeed, we will show convincing empirical evidence supporting this motivation in the following.
        
        \subsection{Graphs of Variance Functions Learned}

\begin{figure*}[t]
	\centering
	\begin{subfigure}{0.192\linewidth}
		\centering
		\includegraphics[width=\linewidth]{{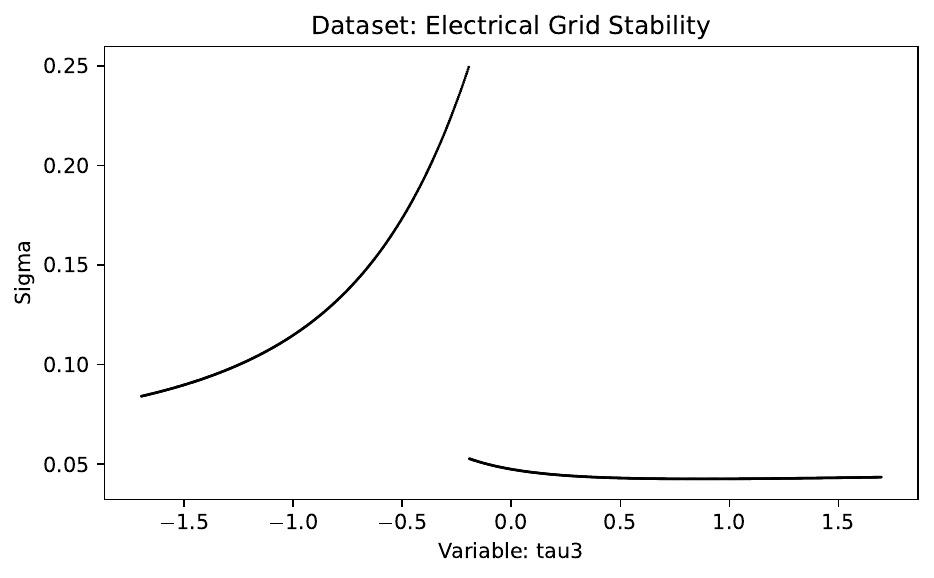}}
		\caption{{\small Electrical}}
	\end{subfigure}
	\centering
	\begin{subfigure}{0.192\linewidth}
		\centering
		\includegraphics[width=\linewidth]{{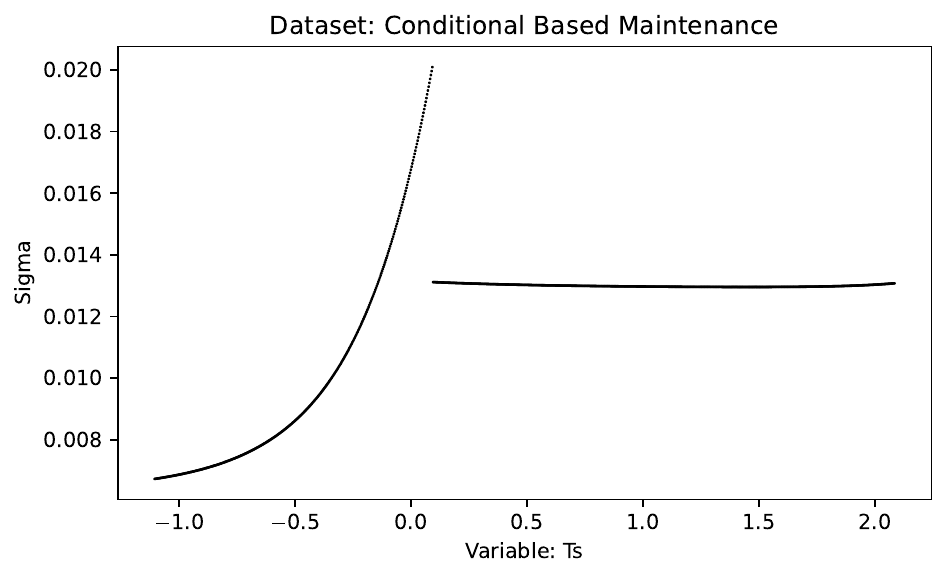}}
		\caption{{\small Conditional}}
	\end{subfigure}
	\centering
	\begin{subfigure}{0.192\linewidth}
		\centering
		\includegraphics[width=\linewidth]{{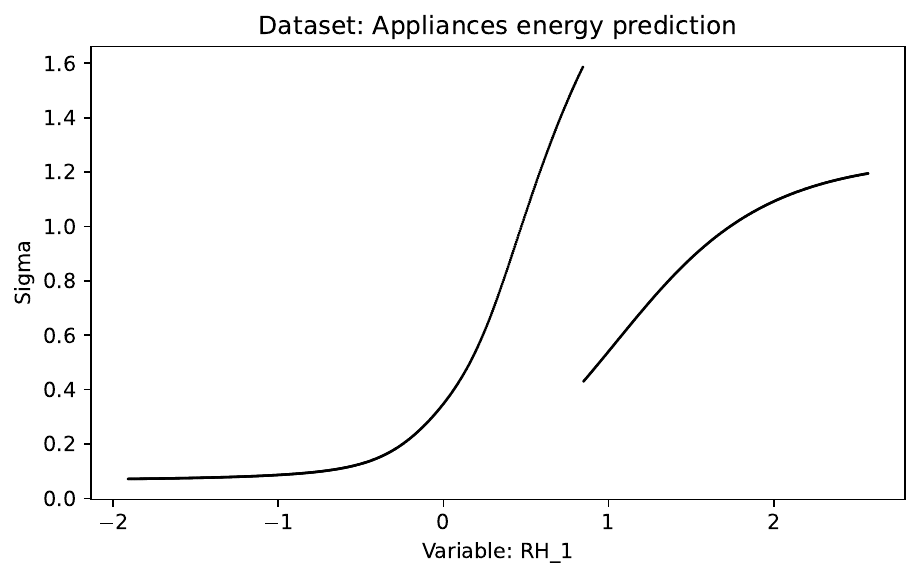}}
		\caption{{\small Appliances}}
	\end{subfigure}
	\centering
	\begin{subfigure}{0.192\linewidth}
		\centering
		\includegraphics[width=\linewidth]{{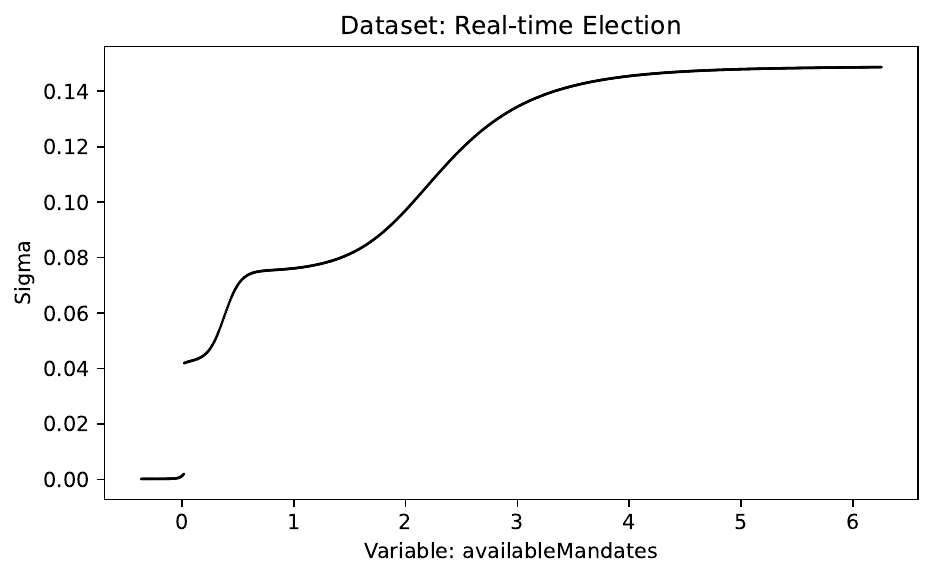}}
		\caption{{\small Real-time}}
	\end{subfigure}
	\centering
	\begin{subfigure}{0.192\linewidth}
		\centering
		\includegraphics[width=\linewidth]{{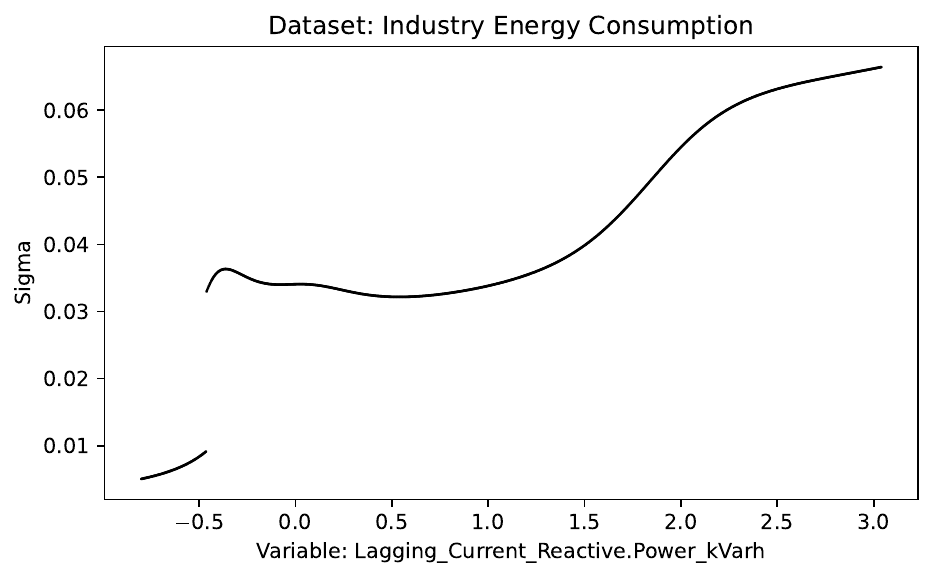}}
		\caption{{\small Industry}}
	\end{subfigure}\\
	\centering
	\begin{subfigure}{0.192\linewidth}
		\centering
		\includegraphics[width=\linewidth]{{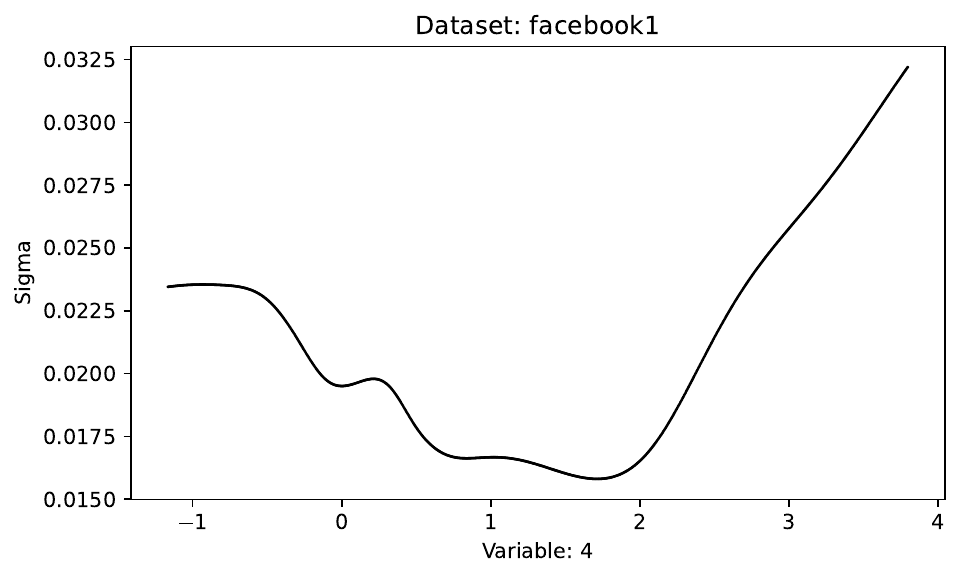}}
		\caption{{\small Facebook1}}
	\end{subfigure}
	\centering
	\begin{subfigure}{0.192\linewidth}
		\centering
		\includegraphics[width=\linewidth]{{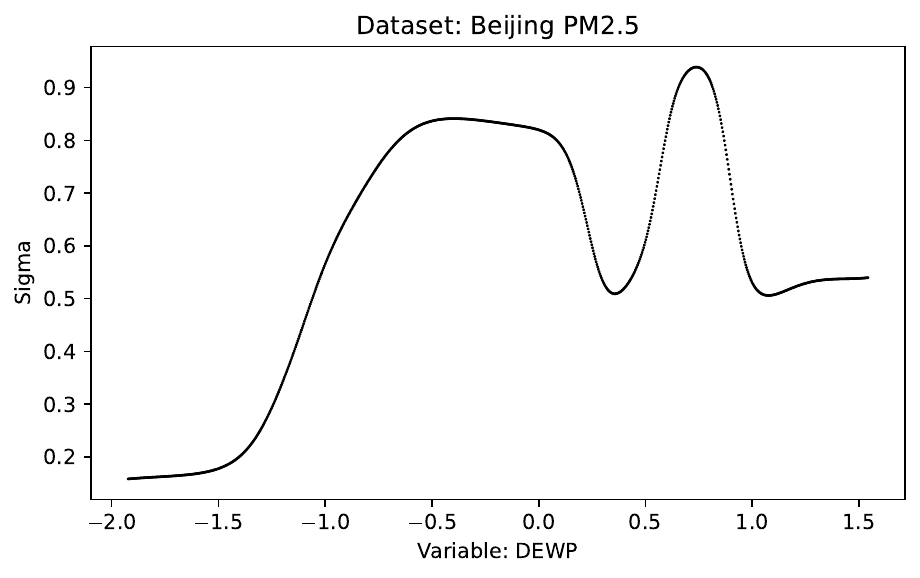}}
		\caption{{\small Beijing}}
	\end{subfigure}
	\centering
	\begin{subfigure}{0.192\linewidth}
		\centering
		\includegraphics[width=\linewidth]{{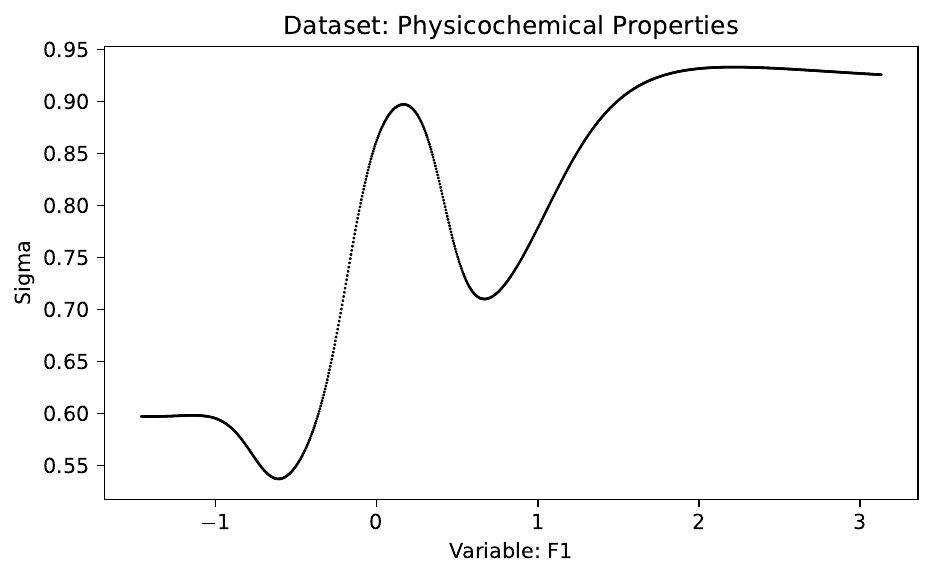}}
		\caption{{\small Physicochemical}}
	\end{subfigure}
	\centering
	\begin{subfigure}{0.192\linewidth}
		\centering
		\includegraphics[width=\linewidth]{{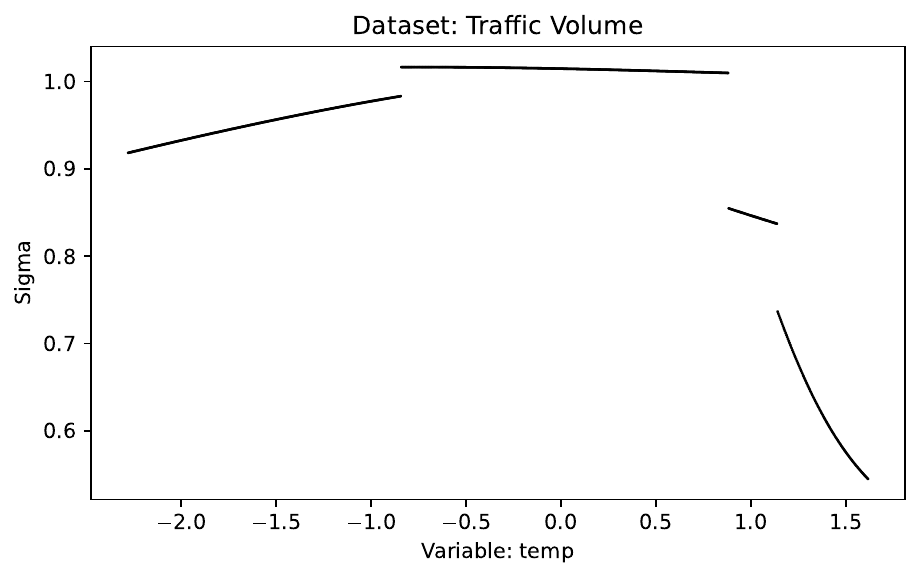}}
		\caption{{\small Traffic}}
	\end{subfigure}
	\centering
	\begin{subfigure}{0.192\linewidth}
		\centering
		\includegraphics[width=\linewidth]{{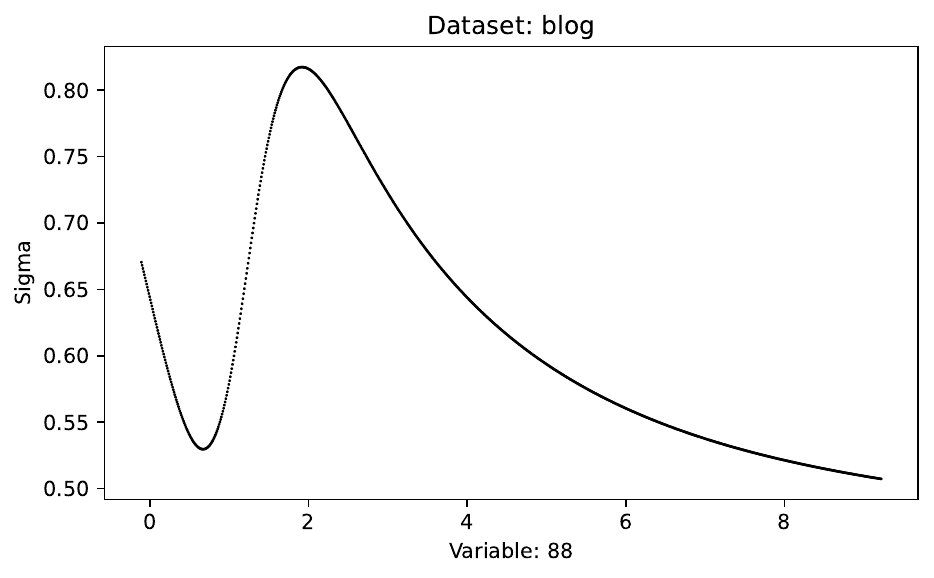}}
		\caption{{\small Blog}}
	\end{subfigure}\\
	\centering
	\begin{subfigure}{0.192\linewidth}
		\centering
		\includegraphics[width=\linewidth]{{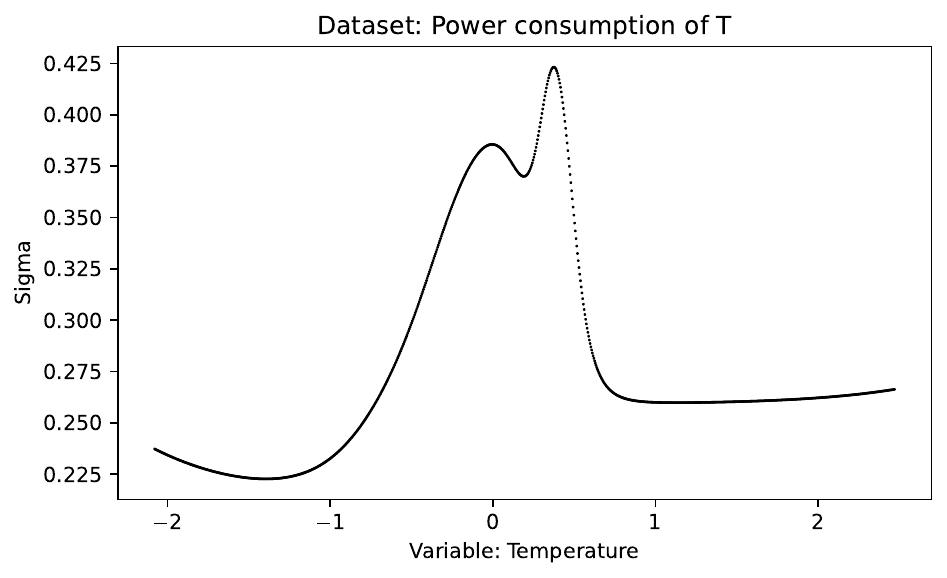}}
		\caption{{\small Power}}
	\end{subfigure}
	\centering
	\begin{subfigure}{0.192\linewidth}
		\centering
		\includegraphics[width=\linewidth]{{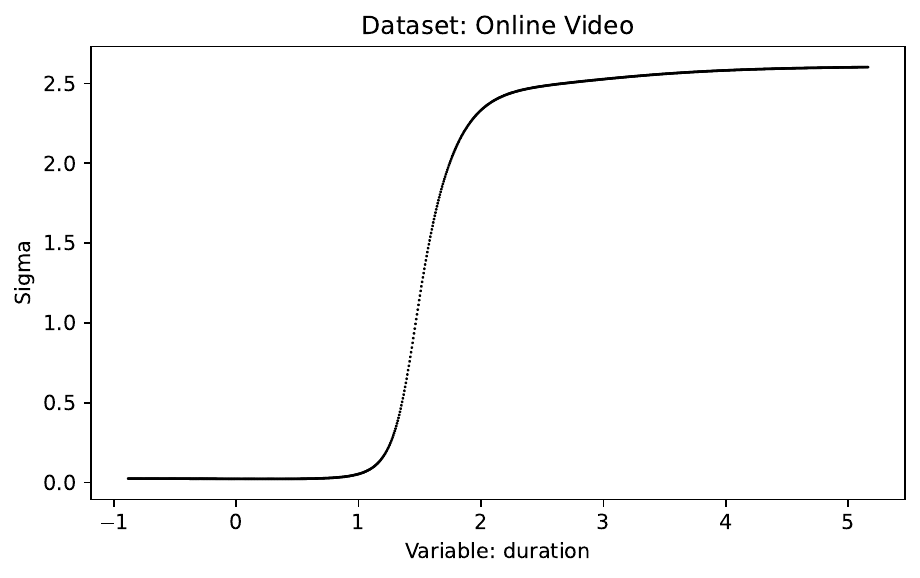}}
		\caption{{\small Online}}
	\end{subfigure}
	\centering
	\begin{subfigure}{0.192\linewidth}
		\centering
		\includegraphics[width=\linewidth]{{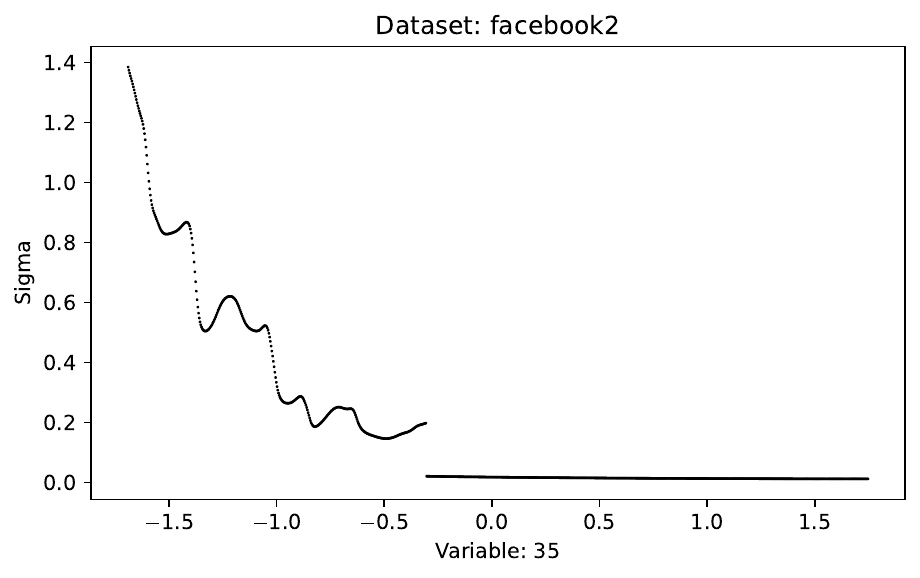}}
		\caption{{\small Facebook2}}
	\end{subfigure}
	\centering
	\begin{subfigure}{0.192\linewidth}
		\centering
		\includegraphics[width=\linewidth]{{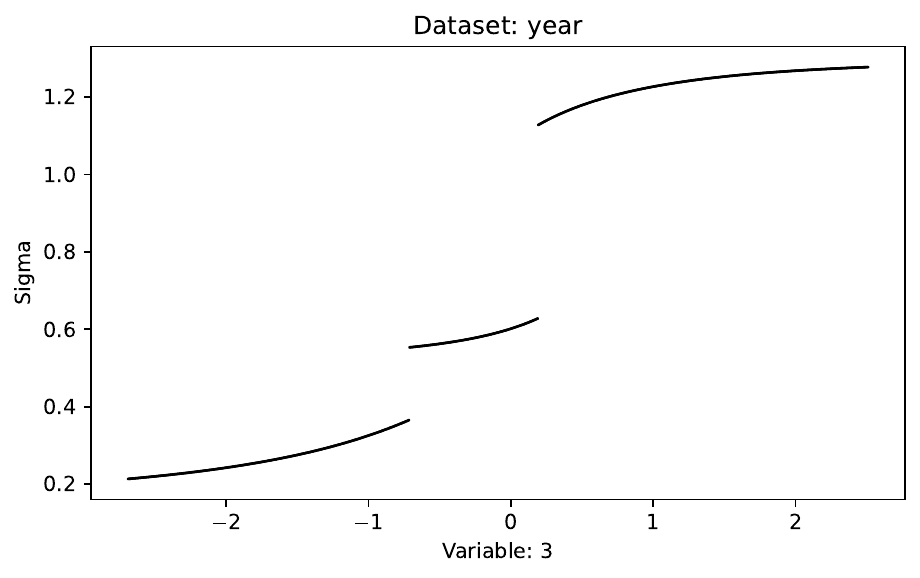}}
		\caption{{\small Year}}
	\end{subfigure}
	\centering
	\begin{subfigure}{0.192\linewidth}
		\centering
		\includegraphics[width=\linewidth]{{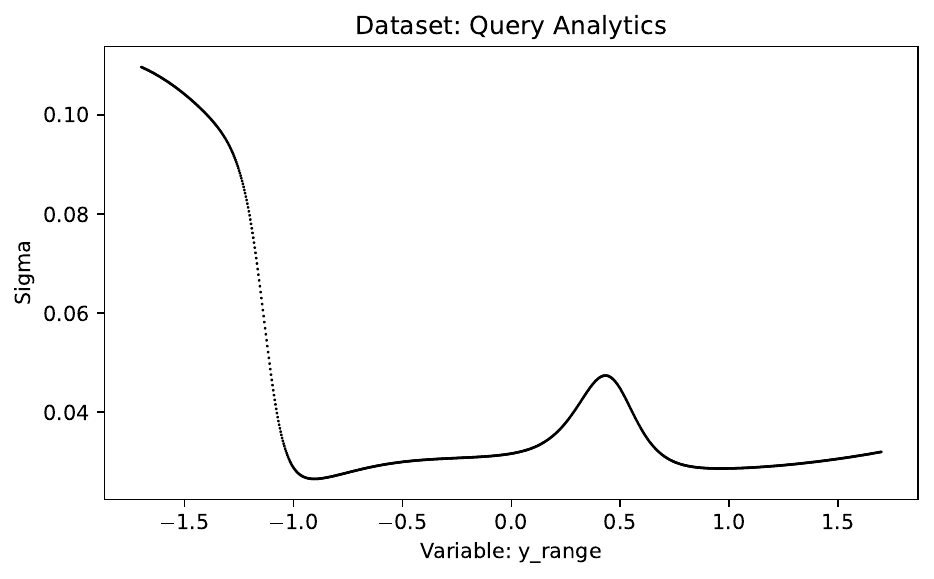}}
		\caption{{\small Query}}
	\end{subfigure}\\
	\centering
	\begin{subfigure}{0.192\linewidth}
		\centering
		\includegraphics[width=\linewidth]{{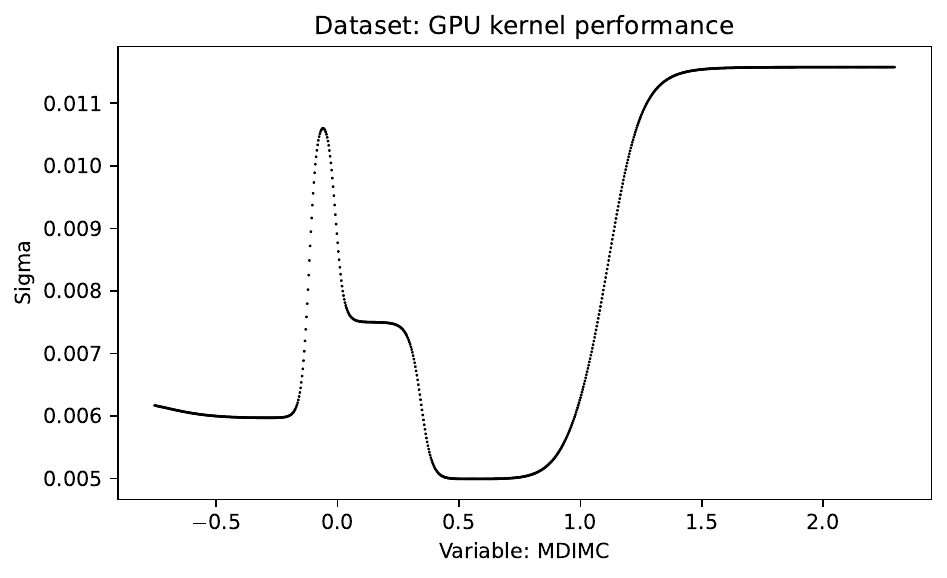}}
		\caption{{\small GPU}}
	\end{subfigure}
	\centering
	\begin{subfigure}{0.192\linewidth}
		\centering
		\includegraphics[width=\linewidth]{{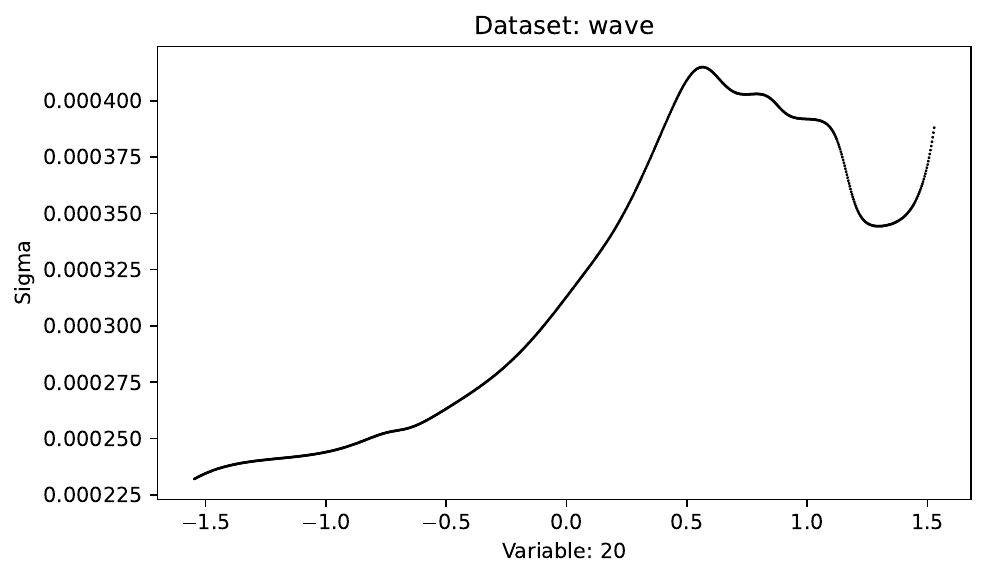}}
		\caption{{\small Wave}}
	\end{subfigure}
	\caption{Graphs of the learned variance functions $\hat{\sigma}(\mathbf{X})$ of USNRT (across all regions). In each dataset, only one feature is selected to the horizontal axis for better visualization. Many datasets exhibit some degrees of heterogeneity: many show discontinuous functions such as Electrical, Conditional, Year, etc.; and some others show different patterns in different segments, such as Facebook1, Facebook2, GPU, etc.}
	\label{usnrt_function_plots}
\end{figure*}
        
        We first plot in Fig. \ref{usnrt_function_plots} the graphs of the learned variance (standard deviation) functions $\hat{\sigma}(\mathbf{X})$ of USNRT (across all regions). In each dataset, only one feature from $\mathbf{X}$ is selected to the horizontal axis for better visualization, while the values of other features are set to their sample means. As we can see, many datasets exhibit some degrees of heterogeneity: many show discontinuous functions such as Electrical, Conditional, Year, etc.; and some others show different patterns in different segments, such as Facebook1, Facebook2, GPU, etc.
        Both the two situations are the reasons for our model to perform better, as discussed in the theoretical section. For the latter one, a split-and-learn strategy will of course reduce the model fitting difficulty.
        
        %%%%%%%%%%%%%%%%%%%%%%%%%%%%%%%%%%%%%%%%%%%%%%%%%%%%%%%%%%%%%%
        
        \subsection{Space Partitioning at Root Node}
        
        \begin{figure*}[t]
                \centering
                \begin{subfigure}[t]{0.31\linewidth}
                        \centering
                        \includegraphics[width=\linewidth]{{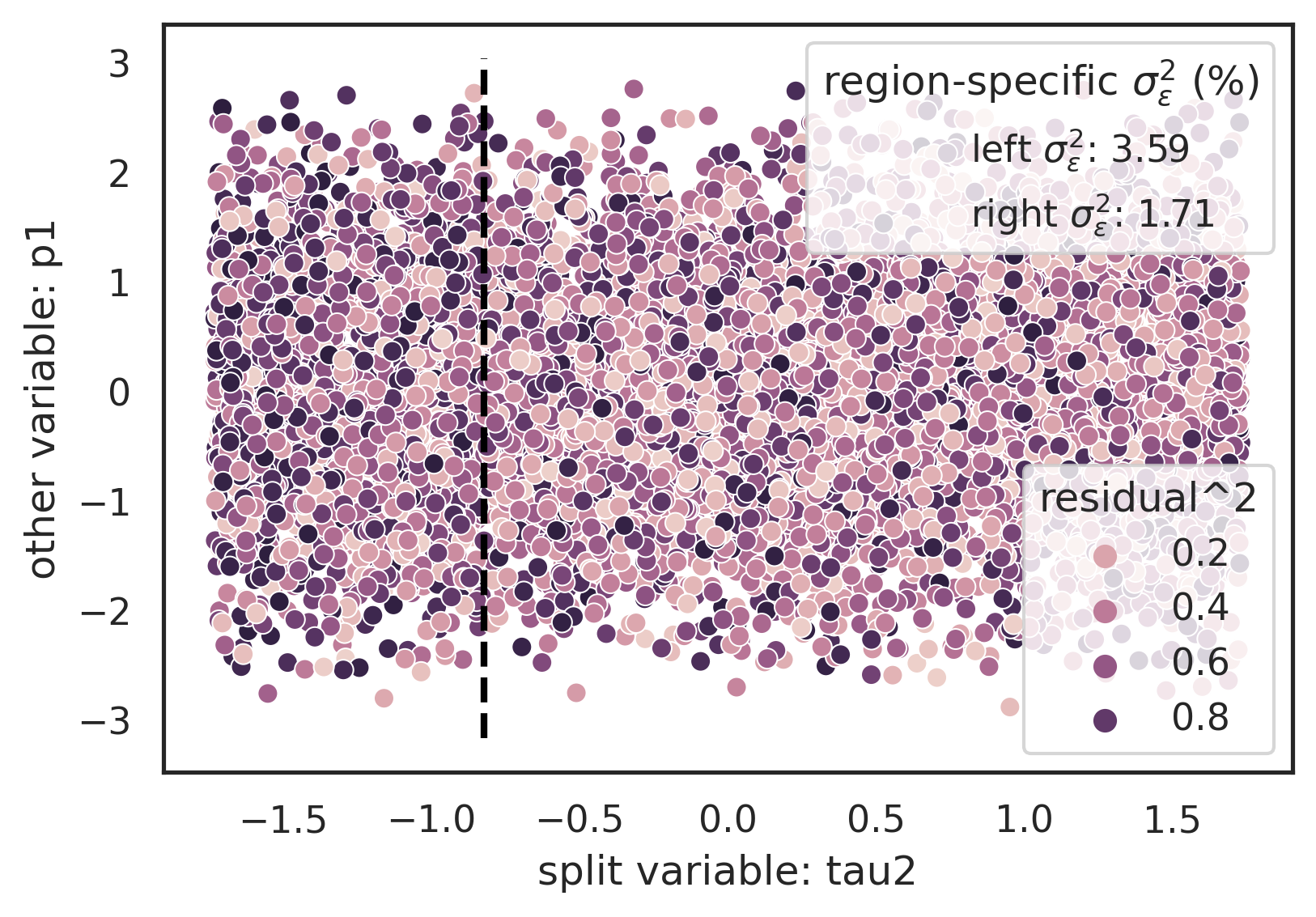}}
                        \caption{{\small Electrical}}
                \end{subfigure}\hspace{0.8em}
                \centering
                \begin{subfigure}[t]{0.31\linewidth}
                        \centering
                        \includegraphics[width=\linewidth]{{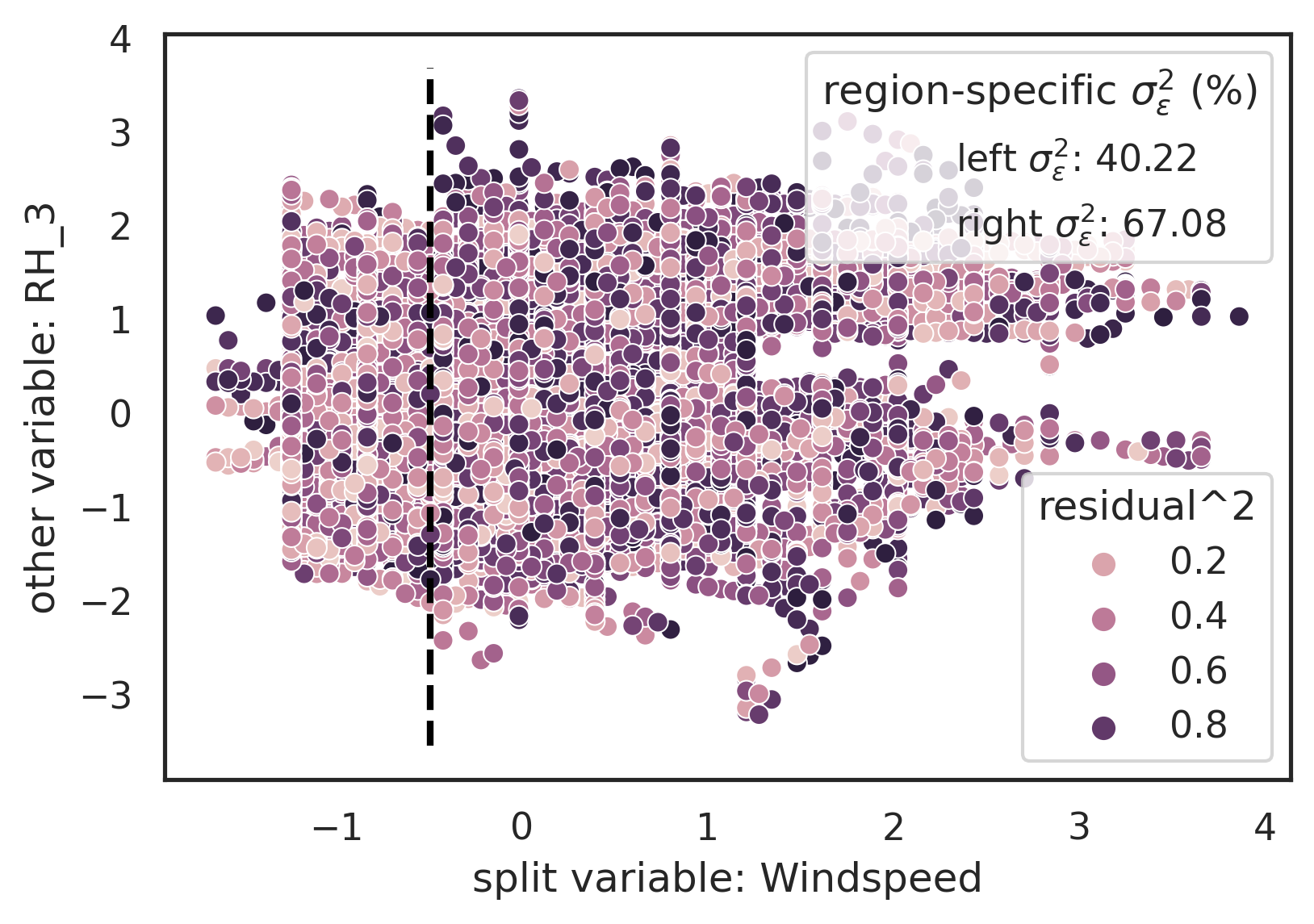}}
                        \caption{{\small Appliances}}
                \end{subfigure}\hspace{0.8em}
                \centering
                \begin{subfigure}[t]{0.31\linewidth}
                        \centering
                        \includegraphics[width=\linewidth]{{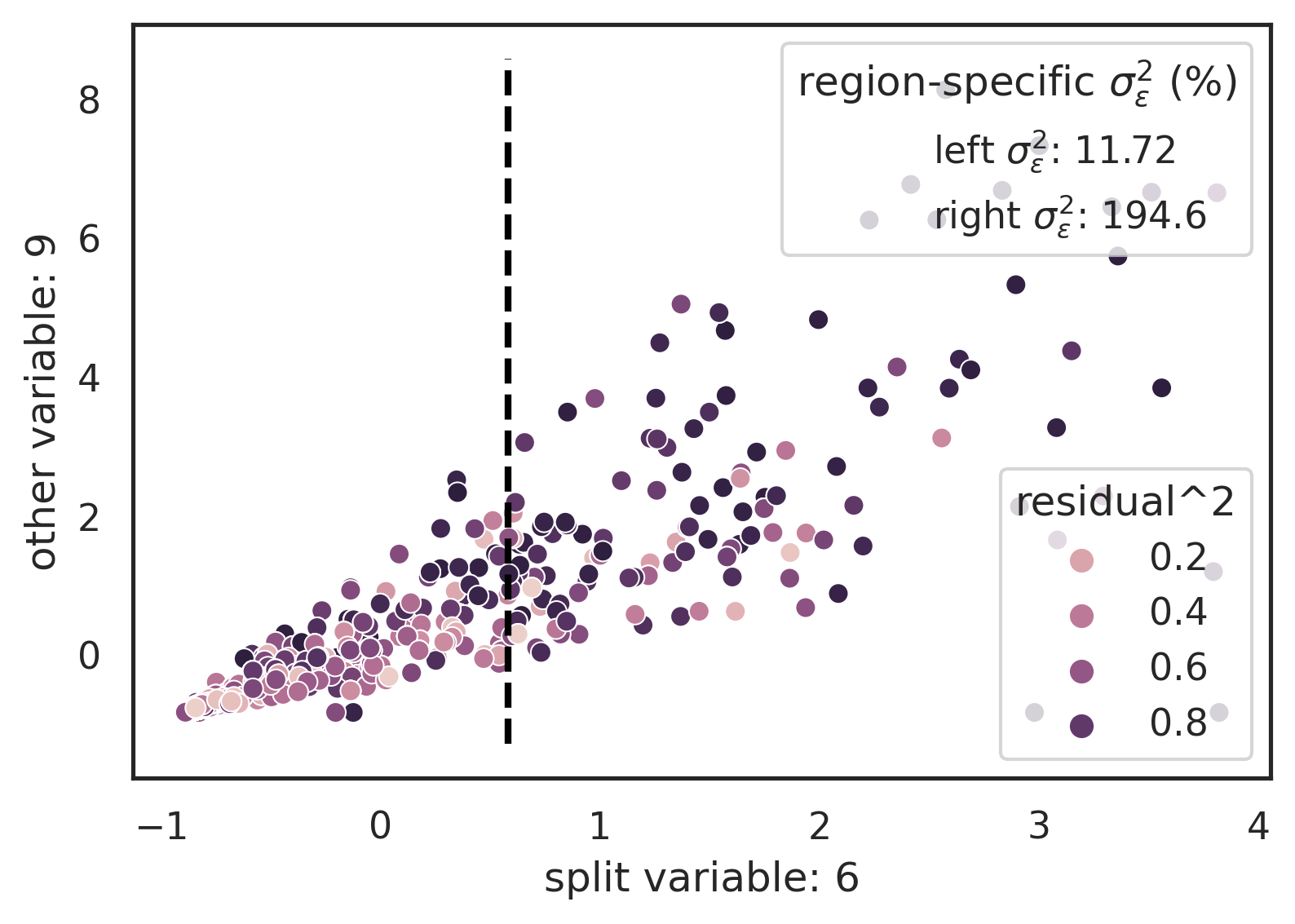}}
                        \caption{{\small Facebook1}}
                \end{subfigure}\\
                \centering
                \begin{subfigure}[t]{0.31\linewidth}
                        \centering
                        \includegraphics[width=\linewidth]{{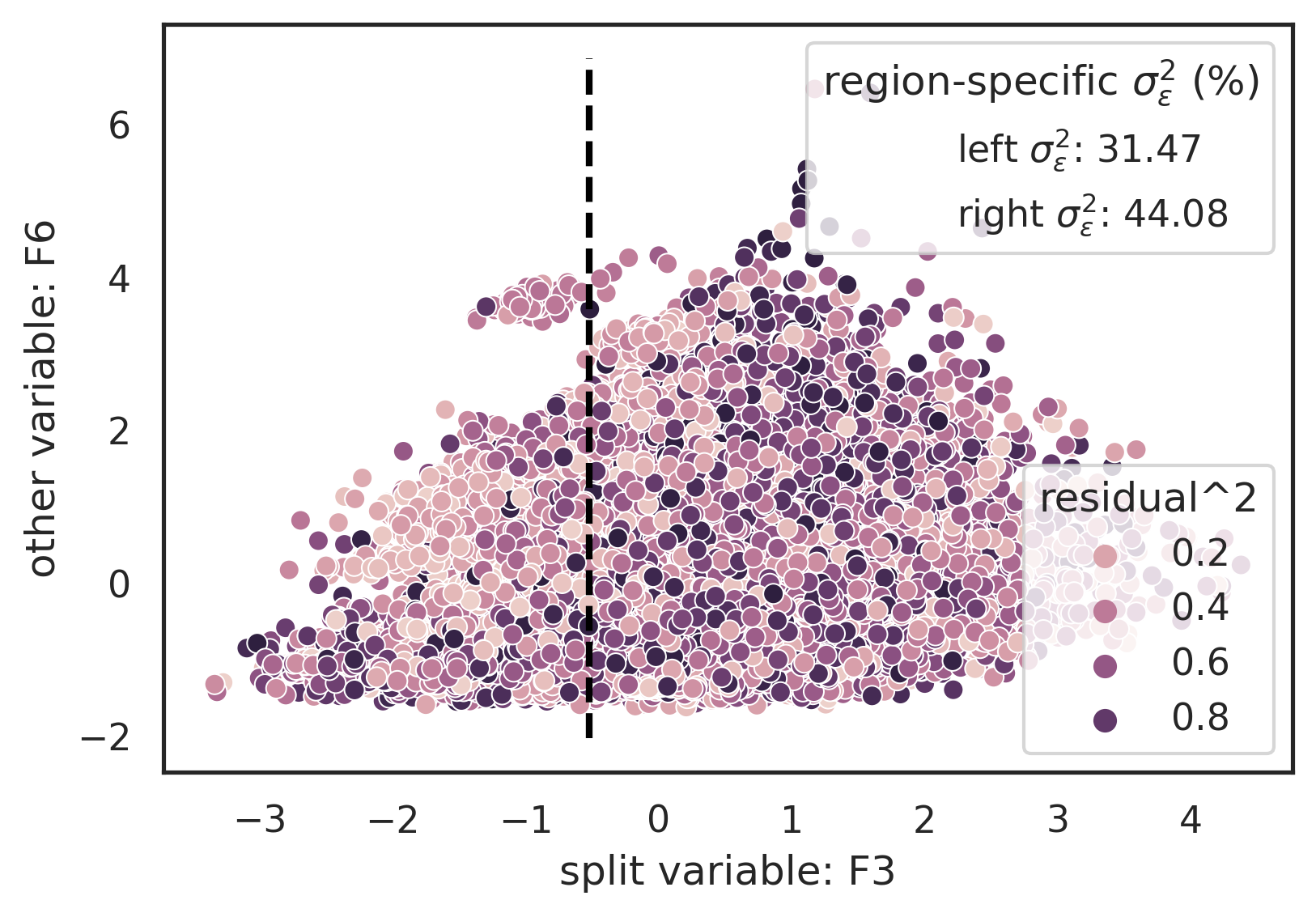}}
                        \caption{{\small Physicochemical}}
                \end{subfigure}\hspace{0.8em}
                \centering
                \begin{subfigure}[t]{0.31\linewidth}
                        \centering
                        \includegraphics[width=\linewidth]{{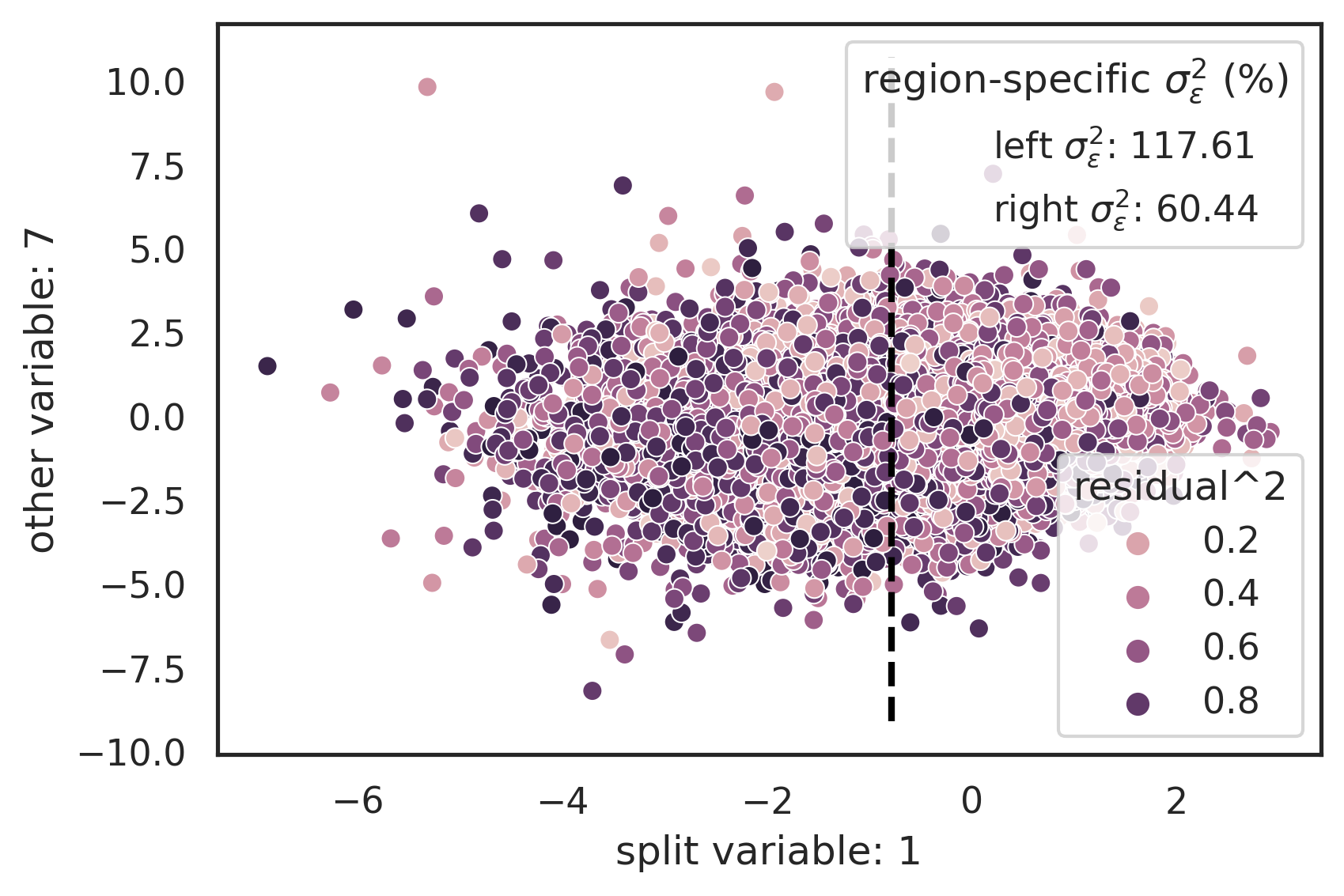}}
                        \caption{{\small Year}}
                \end{subfigure}\hspace{0.8em}
                \centering
                \begin{subfigure}[t]{0.31\linewidth}
                        \centering
                        \includegraphics[width=\linewidth]{{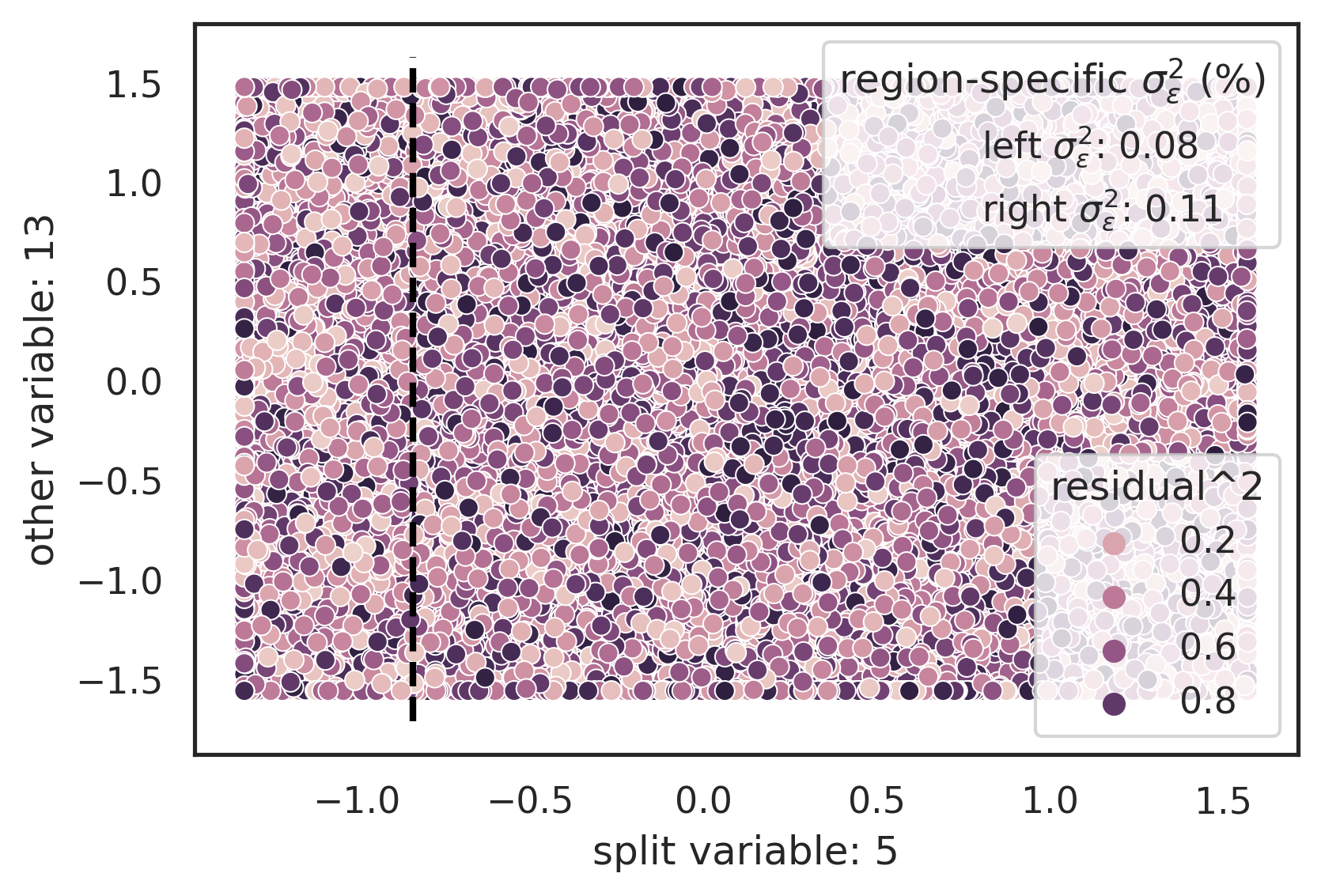}}
                        \caption{{\small Wave}}
                \end{subfigure}
                \caption{Visualization of the partitioning at root node for six datasets: Electrical, Appliances, Facebook1, Physicochemical, Year, and Wave. The whole feature space is projected to two-dimensional. The squared residuals after fitting $\hat{f}_{\text{split}}$ are plotted with different color depths (after a quantile transformation). The dashed line gives the best partition found by USNRT. One can find that the left-region residuals' variance and the right-region residuals' variance are significantly different, especially in datasets Electrical, Facebook1, and Year.}
                \label{split_uncertainty}
        \end{figure*}
        
        Next we visualize how USNRT partitions the feature space, taking the partitioning at the first node or root node as an example. At the root node, USNRT partitions the whole space into two regions.
        Fig. \ref{split_uncertainty} shows the results on 6 datasets: Electrical, Appliances, Facebook1, Physicochemical, Year, and Wave (under one specific training/testing split). The whole feature space is projected to two-dimensional using the split variable and one of other variables. The squared residuals $\left( \mathbf{y}(i)-\hat{f}_{\text{split}}(\mathbf{X}(i)) \right)^2$ for all $i$ are plotted as scattering points with different color depths. The color depths represent the scales of squared residuals after a quantile transformation (to uniform distribution). The dashed line gives the best partition found by USNRT. We also compute the left-region residuals' variance and the right-region residuals' variance (before the quantile transformation) to form a legend. One can find that the left and right variances are significantly different, indicating the possibility of distinct uncertainty natures. The color depths also illustrate such a difference. Such partitioning will be done recursively by USNRT until no heterogeneity can be discovered anymore under some criteria.

        %%%%%%%%%%%%%%%%%%%%%%%%%%%%%%%%%%%%%%%%%%%%%%%%%%%%%%%%%%%%%%
        
        \subsection{Uncertainty Heterogeneity across Leaf Regions}
        
        \begin{figure}[t]
                %\hspace{-2.5em}
                \includegraphics[width=1\linewidth]{{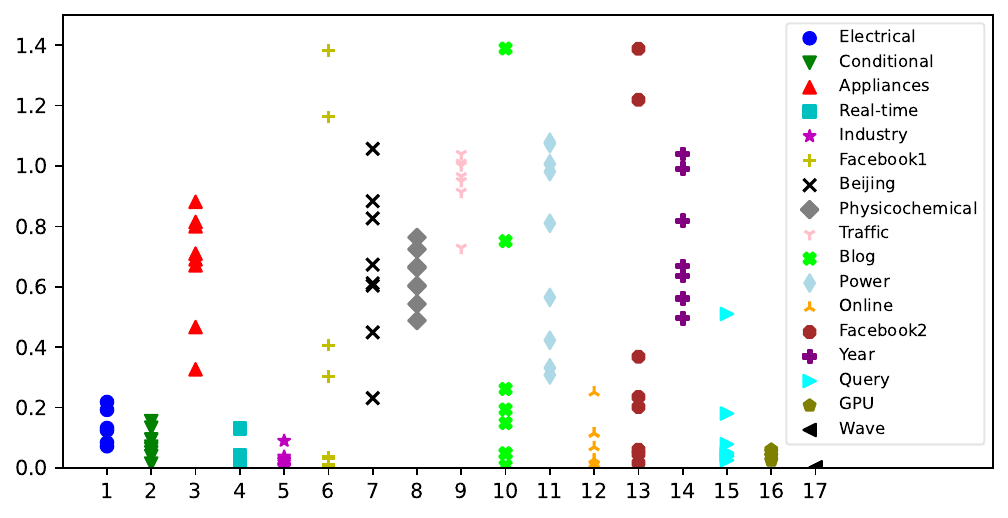}}
                \caption{Leaf region-specific residuals' standard deviations of USNRT. Each type of marker corresponds to one dataset. In many cases, those standard deviations of one dataset are significantly different or heterogeneous, sometimes spreading from 0 to 1.4. This indicates that separate leaf regions have highly distinct uncertainty levels/natures.}
                \label{Uncertainty_difference}
        \end{figure}

                \begin{table}[t]
                \centering
                \caption{\textbf{Uncertainty Heterogeneity Measure}. We define the measure for each dataset by $\text{std}\left(\{\text{log} \hat{\sigma}^{\text{res}}_m\}_{m=1}^M\right)$, where $\hat{\sigma}^{\text{res}}_m$ is the region-$m$ point in Fig. \ref{Uncertainty_difference}. The Percentage Decrease column is from Table \ref{ece_result_table}. The two columns here have a Spearman rank-order correlation 0.44, implying a close relationship between uncertainty heterogeneity and USNRT's performance.}
                \label{uncertainty_heterogeneity_measure}
                %\resizebox{\columnwidth}{!}{
                \footnotesize
                        \begin{tabular}{ccc}
                                \toprule
Dataset & \makecell{Measure\\Sorted} & \makecell{Percentage\\Decrease} \\ \midrule
Facebook1 & 1.80 & 20.50\% \\
Blog & 1.57 & 19.70\% \\
Facebook2 & 1.46 & 20.20\% \\
Online & 1.08 & 53.50\% \\
Real-time & 1.00 & 83.10\% \\
Industry & 0.98 & 27.70\% \\
Query & 0.93 & 30.10\% \\
Wave & 0.92 & 74.60\% \\
Conditional & 0.70 & 63.70\% \\
Power & 0.49 & 2.30\% \\
Beijing & 0.44 & 10.70\% \\
Electrical & 0.40 & -84.90\% \\
Appliances & 0.31 & 35.60\% \\
GPU & 0.31 & 23.50\% \\
Year & 0.26 & 3.90\% \\
Physicochemical & 0.14 & 11.30\% \\
Traffic & 0.11 & -6.60\% \\
 \bottomrule
	                \end{tabular}%}
	        \end{table}

       At last, after learning an USNRT (under one specific training/testing split) on each dataset, we compute the leaf region-specific residuals' variance using all $\left( \mathbf{y}(i)-\hat{f}_m(\mathbf{X}(i)) \right)^2$ when $\mathbf{X}(i)\in\mathcal{R}_m$ and denote the standard deviation as $\hat{\sigma}^{\text{res}}_m$. After obtaining the variances, we then plot the standard deviations as scattering points in Fig. \ref{Uncertainty_difference}. Each type of marker corresponds to one dataset. We can see that in many cases, those standard deviations of leaf regions in one dataset are significantly different or heterogeneous, sometimes spreading from 0 to 1.4. This indicates that separate leaf regions have highly distinct uncertainty levels/natures. This partially reveals the reason for USNRT's success and the difficulty of quantifying uncertainty using one model for the whole feature space. USNRT treats the leaf regions separately after obtaining those leaf regions by tree-structured learning.
       
       It seems that there are some exceptive datasets that have similar uncertainty levels across regions, such as Real-time, Industry, GPU, and Wave. To make the analysis more quantitative, we define a measure of  uncertainty heterogeneity by the standard deviation of the logarithms of all $\hat{\sigma}^{\text{res}}_m$, i.e., $\text{std}\left(\{\text{log} \hat{\sigma}^{\text{res}}_m\}_{m=1}^M\right)$, to measure the discrepancy among the points of one dataset in Fig. \ref{Uncertainty_difference}.
       Then in Table \ref{uncertainty_heterogeneity_measure}, we show this measure of every dataset and a performance indicator of USNRT, i.e., the Percentage Decrease column in Table \ref{ece_result_table}. Now in Table \ref{uncertainty_heterogeneity_measure}, a Spearman rank-order correlation 0.44 between the two columns implies a close relationship between uncertainty heterogeneity and USNRT's performance. These are definitely empirical evidence supporting our motivation of uncertainty heterogeneity.
         %We can also conclude that the uncertainty heterogeneity widely exists in real-world data.
        
        %We also measure and visualize the overall uncertainty level of every leaf region generated by USNRT. Here we simply use the square root of the sample variance of the residuals in each leaf region only to stand for its overall uncertainty level. Fig. \ref{Uncertainty_difference} summarizes the results of 12 UCI datasets (only one split of training and testing sets is used to fix the USNRT structure). In most cases, the uncertainty varies greatly in different regions in the dataset, implying statistically significant difference. It indicates that there is widely existing data heterogeneity in real and large-scale datasets, and discovering such heterogeneity through the effective region partitioning of  USNRT can help improve the deep model performance. Hence, it is strongly necessary to consider data heterogeneity when building deep models.

        \section{Conclusion}
        
                In this paper, we propose a tree-structured local learning model USNRT to recursively partition the feature space and improve uncertainty quantification of variance networks. It is motivated by the widely-existing data heterogeneity in real-world datasets. With the help of a statistical test, we design a splitting procedure to discover the uncertainty heterogeneity across regions in the feature space and construct the tree. Region-specific neural networks are employed to output both the mean and the variance. USNRT has acceptable computational cost and pruning is unnecessary. Experiments on extensive UCI datasets demonstrate that USNRT outperforms some recent popular models on both aleatory uncertainty estimation and total uncertainty estimation. Through visualization, we also show that USNRT can indeed discover uncertainty heterogeneity in the datasets. Moreover, USNRT's performance is stable with respect to the varying neural network structures. It is also easy to determine the most important hyper-parameter: the minimum sample size in leaf nodes (or equivalently, maximum number of leaf nodes).
                
                In the future, it is necessary to take data heterogeneity into consideration for building up better machine learning models, through the techniques of data partitioning like in this paper. It is also interesting to see other forms of combination of deep learning and tree models, although this paper only focuses on uncertainty quantification.% In addition, it is necessary to reduce the computational cost of the proposed USNRT further.
                
                %In this paper, we propose a novel tree-based local neural network model USNRT, to make predictions and quantify uncertainty. USNRT recursively partitions the entire dataset by conducting statistical tests on the uncertainty levels to improve data homogeneity within regions and discover heterogeneity across regions. We integrate neural networks to the internal and leaf nodes of the tree to leverage their modeling ability. We also employ USNRT as the base learner to build up the ensemble version which can enhance the predictive performance. 
        
        % Acknowledgements and Disclosure of Funding should go at the end, before appendices and references
        
        
        %\newpage
        
        \appendix

        \section*{A. The Proof of Proposition \ref{thm:inequality.strict}}
        \label{proof:inequality.strict}
        
        {\it Proof:} Without loss of generality, we assume $f_0$ and $f_1$ are discontinuous at the point $x^*$ on the boundary between $\mathcal{D}_0$ and $\mathcal{D}_1$. In the case where $\sigma_0$ and $\sigma_1$ are discontinuous, the proof is similar.
        Suppose $\min_{f\in\mathcal{F},\sigma\in\mathcal{F}'}\mathbb{E}\left[l\left(\mathbf{y},f\left(\mathbf{X}\right),\sigma(\mathbf{X})\right)\right]$ is solved at $f^*, \sigma^*$, because $f^*$ is a continuous function and $f_0,f_1$ are continuous on $\widebar{\mathcal{D}_0}$ and $\widebar{\mathcal{D}_1}$ respectively, we can find a hypercube $H$ centered at $x^*$ with length $\delta$ such that:
        \begin{align*}
                &  | f^*(x) -f^*(x^*) | <  \epsilon/8, \quad  x\in H, \\
                &  | f_0(x) -f_0(x^*) | <  \epsilon/8, \quad  x\in H\bigcap\mathcal{D}_0, \\
                &  | f_1(x) -f_1(x^*) | <  \epsilon/8, \quad  x\in H\bigcap\mathcal{D}_1,
        \end{align*}
        where $\epsilon=|f_1(x^*)-f_0(x^*)|$. 
        Because $\epsilon\le|f_1(x^*)-f^*(x^*)|+|f^*(x^*)-f_0(x^*)|$, we have $ |f_1(x^*)-f^*(x^*)|\ge \epsilon/2$, or, $|f^*(x^*)-f_0(x^*)|\ge \epsilon/2$.
        Without loss of generality, we assume the latter is true, then
        \begin{align*}
        & \epsilon/2 \le |f^*(x^*)-f_0(x^*)| \le \\
        & |f^*(x^*)-f^*(x)| + |f^*(x)-f_0(x)| + |f_0(x)-f_0(x^*)| \\
        & < \epsilon /8 +|f^*(x)-f_0(x)| + \epsilon / 8, \quad \text{when } x\in H\bigcap\mathcal{D}_0.
        \end{align*}
        So $| f^*(x) -f_0(x) | >  \epsilon/4$ for all $x\in H\bigcap\mathcal{D}_0$. This further implies $ f^*(x) -f_0(x) >  \epsilon/4, ~\forall x\in H\bigcap\mathcal{D}_0$, or, $ f^*(x) -f_0(x) < - \epsilon/4, ~\forall x\in H\bigcap\mathcal{D}_0$. No matter which one is true, we can always successfully construct an $f_2\in\mathcal{F}$ according to the condition 3), such that $f_2(x)=0$ when $x\notin H$ and $0>f_2(x)>-\epsilon /4$ (or $0<f_2(x)<\epsilon /4$) when $x\in H$. Then
        \begin{align*}
        f^*(x)>(f^*+f_2)(x)>f_0(x),\quad x\in H\bigcap\mathcal{D}_0,\\
        (f^*+f_2)(x) = f^*(x),\quad x\in H^c\bigcap\mathcal{D}_0.
        \end{align*}
        The new function $f^*+f_2$ gives: 
        \begin{align*}
        & \mathbb{E}\left[l\left(\mathbf{y}_0,(f^*+f_2)(\mathbf{X}_0),\sigma^*(\mathbf{X}_0)\right)\right] \\
        < & \mathbb{E}\left[l\left(\mathbf{y}_0,f^*(\mathbf{X}_0),\sigma^*(\mathbf{X}_0)\right)\right].
        \end{align*}
        This inequality holds because $\mathbf{X}_0$ and $\mathbf{X}_1$ have positive densities around $x^*$ and $l$ is a proper loss such that a better $f\in\mathcal{F}$ will lead to a smaller expected risk. 
        With this inequality, we have
        \begin{align*}
        & \mathbb{E}\left[l\left(\mathbf{y},f^*(\mathbf{X}),\sigma^*(\mathbf{X})\right)\right] \\
        = ~& \mathbb{E}\left[\mathbb{E}\left[l\left(\mathbf{y},f^*(\mathbf{X}),\sigma^*(\mathbf{X})\right)| Z \right]\right] \\
        = ~& (1-p)\mathbb{E}\left[l\left(\mathbf{y_0},f^*(\mathbf{X_0}),\sigma^*(\mathbf{X_0})\right)\right] \\
        &  + p\mathbb{E}\left[l\left(\mathbf{y_1},f^*(\mathbf{X_1}),\sigma^*(\mathbf{X_1})\right)\right] \\
        > ~& (1-p)\mathbb{E}\left[l\left(\mathbf{y}_0,(f^*+f_2)(\mathbf{X}_0),\sigma^*(\mathbf{X}_0)\right)\right] \\
        &  + p\mathbb{E}\left[l\left(\mathbf{y_1},f^*(\mathbf{X_1}),\sigma^*(\mathbf{X_1})\right)\right] \\
        \ge ~& (1-p)\min_{f\in\mathcal{F},\sigma\in\mathcal{F}'}\mathbb{E}\left[l\left(\mathbf{y}_0,f(\mathbf{X}_0),\sigma(\mathbf{X}_0)\right)\right] \\
        &  + p\min_{f\in\mathcal{F},\sigma\in\mathcal{F}'}\mathbb{E}\left[l\left(\mathbf{y_1},f(\mathbf{X_1}),\sigma(\mathbf{X_1})\right)\right]. \\
        \end{align*}
        Here we can finish the proof.\hfill$\blacksquare$

\end{document}